\documentclass[preprint,12pt,authoryear]{elsarticle}

\usepackage[utf8]{inputenc}
\usepackage[T1]{fontenc}

\usepackage{amsmath}
\usepackage{amssymb}
\usepackage{amsfonts}
\usepackage{booktabs}
\usepackage{multirow}
\usepackage{chngcntr}
\usepackage{nicefrac}
\usepackage{microtype}
\usepackage{xcolor}
\usepackage{url}
\usepackage{float}
\usepackage{placeins}

\usepackage{adjustbox}
\usepackage{hyperref}
\hypersetup{
    colorlinks=false,
    pdfborder={0 0 1},
    linkbordercolor={1 0 0},
    citebordercolor={0 1 0},
    urlbordercolor={0 0 1}
}
\newcommand{\figref}[1]{\hyperref[#1]{Figure~\ref*{#1}}}
\newcommand{\tabref}[1]{\hyperref[#1]{Table~\ref*{#1}}}
\journal{Expert Systems with Applications}

\begin{document}

\begin{frontmatter}

\title{Evaluating Sequence Assembly Strategies for Differentially Private Synthetic Time-Series Forecasting}

%% Double-anonymized manuscript for ESWA peer review.
%% Author names, affiliations, and contact information are omitted.

\author[aff1]{Guoxiong Long}
\ead{longguoxiong@stu.mju.edu.cn}

\author[aff1]{Huizhen Huang}
\ead{huanghuizhen@stu.mju.edu.cn}

\author[aff1]{Qikun Cai}
\ead{caiqikun@stu.mju.edu.cn}

\author[aff1]{Tao Huang\corref{cor1}}
\ead{huang-tao@mju.edu.cn}

\author[aff1]{Chen Hou\corref{cor1}}
\ead{houchen@mju.edu.cn}

\cortext[cor1]{Corresponding authors}

\affiliation[aff1]{
    organization={School of Computer and Big Data, Minjiang University},
    addressline={No. 200 Xiyuangong Road},
    city={Fuzhou},
    postcode={350108},
    state={Fujian},
    country={China}
}
\begin{abstract}
Differentially private time-series generators commonly produce fixed-length synthetic windows, whereas downstream forecasting models often require long continuous training sequences. How these windows are assembled after generation can therefore alter the effective synthetic data presented to a forecaster. We study this post-generation sequence assembly process by systematically varying overlap rates and window-weighting schemes and evaluating the resulting sequences in terms of boundary continuity, statistical and temporal fidelity, and Train-on-Synthetic-Test-on-Real (TSTR) forecasting utility. Across four public datasets and five forecasting models, the results reveal a clear forecaster-dependent assembly principle: downstream TSTR utility is jointly shaped by the forecaster, overlap rate, and window-weighting schemes, leading to distinct assembly preferences across forecasting models. Increased overlap generally improves boundary continuity, but improvements in continuity or individual fidelity diagnostics do not consistently reduce forecasting error, indicating that these diagnostics alone are insufficient for selecting assembly configurations. Complete five-forecaster assembly grids, together with matched Train-on-Real-Test-on-Real (TRTR) references, further characterize these regularities and quantify assembly-dependent utility relative to real-data training. We then validate the identified principles in a domain-specific operational electricity-load forecasting application, where forecaster-specific sensitivity to sequence assembly persists and provides practical guidance for prioritizing candidate configurations for downstream evaluation. Overall, the findings establish post-generation sequence assembly as a forecaster-aware model-selection problem and provide actionable guidance for deploying differentially private synthetic time series in forecasting systems.
\end{abstract}

%% Graphical abstract is optional. Uncomment and provide a file if needed.
%% \begin{graphicalabstract}
%% \includegraphics[width=\textwidth]{graphical_abstract.pdf}
%% \end{graphicalabstract}

%% Research highlights (also supplied as highlights.txt in this project).
%% Research highlights are supplied separately in highlights.txt.
%% \begin{highlights}
%% \item Sequence assembly is evaluated for synthetic windows from DP-SGD-trained generators.
%% \item Continuity and fidelity metrics do not uniformly predict TSTR utility.
%% \item Matched TRTR provides a real-training reference for synthetic forecasting utility.
%% \item Multi-dataset results inform forecaster-aware assembly candidate priorities.
%% \item Three DP-ConvVAE settings provide a separate cross-generator check.
%% \end{highlights}


\end{frontmatter}

\section{Introduction}

Differentially private time-series generators, commonly trained with
differentially private stochastic gradient descent (DP-SGD), can produce
synthetic data for downstream forecasting while providing formal privacy
guarantees during model training
~\citep{dwork2014algorithmic,abadi2016deep,jordon2018pate}.
Many generators emit fixed-length synthetic windows, whereas downstream
forecasting models are typically trained on long, continuous sequences.
The generated windows must therefore be combined after generation before they
can serve as forecasting data. We refer to this operation as
\emph{post-generation sequence assembly}. Existing work on
privacy-preserving synthetic time series has focused primarily on private
generator training, statistical fidelity, and downstream utility
~\citep{frigerio2019differentially,dockhorn2022differentially,
lamp2023glucosynth,jiang2024fedstdg}. The assembly stage has received much
less systematic attention, even though different overlap rates and
window-weighting schemes can change the effective training sequence presented
to a downstream forecaster while the trained generator remains fixed.

\begin{figure}[t]
    \centering
    \includegraphics[width=0.65\textwidth]{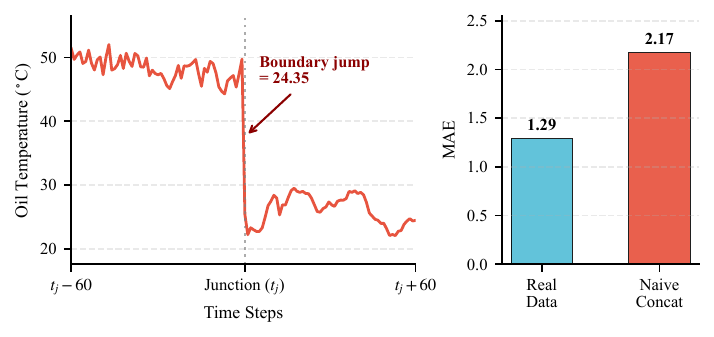}

    \caption{Illustration of post-generation sequence assembly on ETTh1.
    \textbf{Left:} direct concatenation of independently generated windows
    creates a visible boundary discontinuity.
    \textbf{Right:} forecasting performance of the real-data training
    reference \textbf{(blue)} under Train-on-Real-Test-on-Real (TRTR)
    evaluation and the naive-concatenation synthetic-data baseline
    \textbf{(red)} under Train-on-Synthetic-Test-on-Real (TSTR) evaluation.}

    \label{fig:teaser}
\end{figure}

Sequence assembly is therefore a consequential design stage rather than a
formatting operation. Direct concatenation of independently generated windows
can introduce discontinuities at segment junctions, whereas overlap and
window weighting can reduce these local artifacts. At the same time, the
assembly operation can alter marginal distributions, temporal dependence, and
the temporal patterns available to a forecasting model. Consequently, the
smoothest assembled sequence need not be the most useful one for forecasting.
\figref{fig:teaser} illustrates this issue on ETTh1: direct concatenation
creates a visible boundary discontinuity, and a forecaster trained on the
resulting synthetic sequence exhibits a different error level from its matched
real-data training reference.

We evaluate sequence assembly along three complementary dimensions: boundary
continuity, statistical and temporal fidelity, and downstream forecasting
utility. Forecasting utility is measured with Train-on-Synthetic-Test-on-Real
(TSTR), in which a forecaster is trained on an assembled synthetic sequence
and evaluated on held-out real observations. Matched
Train-on-Real-Test-on-Real (TRTR) results provide the corresponding
real-training reference. This joint view allows us to determine which
properties of an assembled sequence are informative for downstream utility
and which assembly choices should be prioritized for a specific forecaster.

Our study follows a discovery-to-application structure. We first
systematically vary overlap rates and window-weighting schemes on ETTh1, ETTm1,
Weather, and Appliances while holding generator checkpoints fixed. Complete
assembly grids across LSTM, RNN, LightGBM, ARIMA, and Prophet reveal a stable
forecaster-dependent regularity: recurrent models more often favor
intermediate-to-high overlap, ARIMA and Prophet favor zero or low overlap,
and LightGBM is more dataset dependent. Improvements in boundary continuity
or an individual fidelity diagnostic do not consistently reduce TSTR error,
so assembly candidates must be evaluated with the intended forecaster. We then validate these forecaster-dependent regularities in an operational electricity-load forecasting task, where the same broad forecaster-dependent assembly pattern persists.

The main contributions of this work are as follows:
\begin{itemize}

    \item We formulate post-generation sequence assembly as an explicit design stage for differentially private synthetic time-series forecasting and systematically evaluate the effects of overlap rates and window-weighting schemes while holding the trained generator fixed.

    \item Across four public datasets and five forecasters, we identify forecaster-dependent assembly regularities and show that improvements in boundary continuity or individual fidelity diagnostics do not consistently translate into higher TSTR forecasting utility.

    \item We utilize matched TRTR references to contextualize forecasting utility relative to real-data training and validate the forecaster-dependent assembly regularities in an operational electricity-load application. Moreover, we provide practical guidance for candidate prioritization in downstream evaluation.

\end{itemize}

\section{Related Work}

\subsection{Privacy-Preserving Time-Series Synthesis and Forecasting Utility}

Differential privacy provides a formal framework for learning from
sensitive data while limiting the influence of individual records on the
released model or output~\citep{dwork2014algorithmic}. In private machine
learning, differentially private stochastic gradient descent (DP-SGD) and
R\'enyi differential privacy (RDP) accounting are widely used to train and
analyze data-driven models
~\citep{abadi2016deep,mironov2017renyi}. Earlier private synthetic-data
methods include teacher--student and GAN-based mechanisms
~\citep{jordon2018pate,frigerio2019differentially}, while differentially
private diffusion training provides an alternative generative approach
~\citep{dockhorn2022differentially}. In parallel, diffusion models have
been applied to several time-series tasks, including imputation,
forecasting, and generation
~\citep{tashiro2021csdi,yuan2024diffusion,fan2024mgtsd,
kollovieh2023predict,yang2026survey}.

Recent studies have also considered downstream task utility when
evaluating privacy-preserving synthetic time series. GlucoSynth develops
a differentially private generator for glucose traces and evaluates
forecasting utility using the Train-on-Synthetic-Test-on-Real (TSTR)
paradigm~\citep{lamp2023glucosynth}. FedSTDG combines local differential
privacy with federated learning and a recurrent Wasserstein autoencoder,
with utility assessed through downstream prediction tasks
~\citep{jiang2024fedstdg}. \citet{bale2026protect} examine the relationship between
privacy protection and forecasting accuracy across several protection
mechanisms, including differential privacy. Together, these studies illustrate that the
evaluation of privacy-preserving time-series data involves not only
privacy guarantees and statistical similarity, but also the extent to
which protected or synthetic data remain useful for downstream
forecasting.

Recent work has further broadened the design space for
privacy-preserving time-series synthesis. PETS introduces an
inference-time differentially private synthesis approach in which private
data guide synthetic-output selection at inference time
~\citep{luo2026pets}. PrivLoad develops a diffusion-based
privacy-preserving method for synthesizing smart-grid load profiles
~\citep{yuan2025privload}. These studies primarily intervene at the
generator, training, or synthesis stage. Our study instead holds the
trained private generator fixed and examines a subsequent stage: how
independently generated synthetic windows are assembled into the
continuous sequence supplied to a forecasting model.

The distinction between generator training and post-generation assembly
is also relevant to the interpretation of privacy in temporal pipelines.
Privacy guarantees depend on the protected unit and on how temporal
observations are organized into training examples. \citet{schuchardt2025structured} study structured temporal subsampling in
forecasting pipelines and develop privacy analyses for event- and user-level
guarantees under such sampling. Their analysis highlights the
importance of explicitly defining the privacy unit when temporal records
are derived from structured sequences. In our setting, the private generator operates on fixed-length temporal windows,
whereas the continuous synthetic sequence used for downstream forecasting is
assembled only after generation. Generator training and sequence assembly
therefore occupy different stages of the privacy-preserving pipeline. We
explicitly distinguish the privacy unit used for generator training from the
subsequent assembly operations on released quantities, which constitute
differential-privacy post-processing.

\subsection{Synthetic Time-Series Evaluation and Post-Generation Sequence Assembly}

Synthetic time-series quality is commonly evaluated from multiple
perspectives. Existing surveys emphasize that distributional, temporal,
and task-based metrics capture different properties of synthetic data
~\citep{stenger2024evaluation}. Recent benchmarking work has further
emphasized multidimensional evaluation of time-series generators.
ConTSG-Bench evaluates conditional time-series generation using
complementary criteria including generation fidelity, condition
adherence, and downstream task utility~\citep{lan2026contsgbench}.
Distributional measures such as Wasserstein
distance~\citep{peyre2019computational} quantify differences between
real and synthetic value distributions, while autocorrelation and
related statistics characterize temporal dependence. Task-based
evaluation adds a direct measure of whether synthetic data preserve
information that is useful to a downstream model. In particular, TSTR
trains a model on synthetic data and evaluates it on held-out real
observations~\citep{jordon2018pate,yoon2019time,lamp2023glucosynth}.

These criteria describe complementary aspects of synthetic-data quality.
A sequence-assembly configuration can alter local transitions, marginal
distributions, temporal dependence, and the temporal structures
available to a forecaster. Consequently, improvement under one
diagnostic does not by itself imply higher downstream forecasting
utility. This motivates evaluating post-generation sequence assembly
jointly through local boundary continuity, statistical and temporal
fidelity, and task-based forecasting utility.

The mechanics of sequence assembly are related to established
signal-processing methods. Overlap-add reconstruction and tapering
windows such as Hann and triangular windows are commonly used when
combining adjacent signal blocks~\citep{oppenheim2010discrete}. In the
present setting, however, the input blocks are independently generated
synthetic windows, and the objective is to construct a continuous
synthetic training sequence for a downstream forecaster. We therefore
use \emph{post-generation sequence assembly} for the operation studied
here, while retaining \emph{overlap-add reconstruction} when referring
to the established signal-processing technique. This distinction
separates reconstruction of an underlying signal from construction of a
synthetic sequence for downstream learning.

Taken together, prior work provides methods for privacy-preserving
time-series synthesis, multidimensional evaluation of synthetic data,
and overlap-based signal combination, but the post-generation assembly
stage remains comparatively underexplored in downstream forecasting
pipelines. In particular, when the trained private generator is held
fixed, it remains unclear how different overlap rates and
window-weighting schemes alter the effective synthetic training
sequence, whether improvements in continuity or individual fidelity
diagnostics consistently correspond to improvements in forecasting
utility, and whether preferred assembly configurations differ across
downstream forecasters.

To address these questions, we systematically vary overlap rates and
window-weighting schemes under fixed generator checkpoints and evaluate
the resulting sequences in terms of boundary continuity, statistical and
temporal fidelity, and TSTR forecasting utility. The multi-dataset,
five-forecaster analysis is used to identify recurring
forecaster-dependent assembly regularities. These regularities are then
examined in an operational electricity-load forecasting application,
while matched Train-on-Real-Test-on-Real (TRTR) references provide
real-training context and a second generator architecture provides a
separate cross-generator check.

\section{Methodology}
\label{sec:method}
The methodology formalizes post-generation sequence assembly and links two analytical propositions to the joint evaluation of continuity, fidelity, and forecasting utility. We first define the synthesis-and-assembly pipeline and
its privacy scope, then characterize the local transition mechanism and the
sequence properties that a boundary diagnostic does not capture.

\subsection{Problem Formulation and Evaluation Scope}
\label{sec:problem}
\subsubsection{Segmented Synthesis and Forecasting Tasks}
\label{sec:problem_formulation}

Let $\mathcal{D}=\{z_t\}_{t=1}^{T_{\mathrm{train}}}$ denote the raw
time-series observations in the training interval. Preprocessing $P$,
including MinMax normalization fitted on this interval, produces
$\mathcal{D}_{\mathrm{norm}}=P(\mathcal{D})$. A deterministic
window-extraction map $W$ then constructs the normalized window dataset
\begin{equation}
    \mathcal{D}_{\mathrm{win}}=W(\mathcal{D}_{\mathrm{norm}})
    =\{X_m\}_{m=1}^{N},
    \qquad
    X_m\in\mathbb{R}^{L\times d},
    \label{eq:raw_to_window_dataset}
\end{equation}
where $L$ is the window length and $d$ is the number of modeled variables.
The generator is trained on these normalized fixed-length window records.
The extraction procedure is fixed throughout the study and
is distinct from the placement stride $k$, which is varied only after
generation. Normalization and window-extraction settings are
reported in \hyperref[app:implementation]{Appendix~A}. The main experiments
evaluate one target channel per dataset; the corresponding targets and
preprocessing protocols are specified in the experimental protocol.

A generative model is trained with DP-SGD on
$\mathcal{D}_{\mathrm{win}}$ and produces synthetic windows
\begin{equation}
    S_i\in\mathbb{R}^{L\times d},
    \qquad i=0,\ldots,M-1.
\end{equation}
These windows are subsequently assembled into a continuous synthetic
sequence by an assembly operator $A$,
\begin{equation}
    \tilde{S}
    =
    A(S_0,\ldots,S_{M-1}).
    \label{eq:assembly_general}
\end{equation}

For placement stride $k$, window $S_i$ is assigned to global offset
$ik$ and covers the interval associated with that position on the target
timeline. Calendar or positional conditioning variables are assigned
consistently with this timeline. For $M$ windows of common length $L$,
the pre-crop assembled length is
\begin{equation}
    T_{\mathrm{raw}}
    =
    L+(M-1)k.
    \label{eq:assembled_length}
\end{equation}
The sequence-assembly stage therefore combines generated windows in their
prescribed temporal order.

The assembled synthetic sequence is then used to train downstream
forecasting models. Under the Train-on-Synthetic-Test-on-Real (TSTR)
protocol, a forecaster is trained on the assembled synthetic data and
evaluated on held-out real observations. Because assembly occurs after
generator training, it can alter the effective training sequence seen by a
forecaster while leaving the DP-SGD-trained generator unchanged.

\subsubsection{Privacy Scope of the Synthesis-and-Assembly Pipeline}
\label{sec:privacy_scope}

The primary generator is trained using DP-SGD with per-example gradient
clipping and Gaussian noise injection~\citep{abadi2016deep}. Each normalized
fixed-length window in $\mathcal{D}_{\mathrm{win}}$ is treated as one DP-SGD
example. Privacy loss is tracked using a R\'enyi Differential Privacy (RDP)
accountant~\citep{mironov2017renyi}, yielding
$(\epsilon_{\mathrm{train}},\delta_{\mathrm{train}})
=(9,10^{-5})$ under the window-record neighboring relation used for
generator training.

A separate differentially private calibration mechanism is applied to the
original target observations under the fixed timestamp-grid observation
adjacency, with
$(\epsilon_{\mathrm{cal}},\delta_{\mathrm{cal}})
=(1,10^{-6})$. The two mechanisms are reported under their respective
privacy definitions.

Once the synthetic windows and noisy calibration statistics have been
released, the sequence-assembly operations examined in this study---including
calibration-based adjustment, smoothing of noisy calibration statistics,
overlap averaging, and window weighting---depend only on released quantities
and fixed hyperparameters. They therefore constitute differential-privacy
post-processing and introduce no additional privacy loss. Detailed privacy-accounting and
calibration settings are provided in
\hyperref[app:implementation]{Appendix~A} and
\hyperref[app:dp_details]{Appendix~B}.

\subsubsection{Evaluation Objectives}
\label{sec:evaluation_objectives}

The evaluation examines how assembly configurations affect boundary
continuity, statistical and temporal fidelity, and downstream forecasting
utility while the trained generator is held fixed. ETTh1, ETTm1, Weather, and
Appliances form the public-data discovery set used to identify
forecaster-dependent assembly principles. Matched Train-on-Real-Test-on-Real
(TRTR) references place the corresponding TSTR results on the real-training
error scale. The operational electricity-load dataset is then used to apply
and validate the public-dataset candidate priorities in a domain-specific
forecasting task.

\subsection{Conditional Boundary-Junction Reduction}
\label{sec:diagnostic_properties}
\label{sec:principle_boundary}
For window length $L$ and placement stride $k$, the overlap rate is
\begin{equation}
\rho=(L-k)/L.
\label{eq:overlap_rate}
\end{equation}
Let $\hat{S}_i$ denote the segment supplied to the assembly operator after
any calibration used by the corresponding experimental configuration; when no calibration is applied,
$\hat{S}_i=S_i$. For global time index $t$, define the local index
\begin{equation}
    u_i(t)=t-ik
\end{equation}
and the coverage indicator
\begin{equation}
    \mathbf{1}_{i,t}
    =
    \mathbf{1}\{1\leq u_i(t)\leq L\}.
\end{equation}

The normalized overlap-add assembly is
\begin{equation}
    \tilde{S}(t)
    =
    \frac{
        \sum_i
        \mathbf{1}_{i,t}
        w(u_i(t))
        \hat{S}_i(u_i(t))
    }{
        \max\left\{
            \sum_i
            \mathbf{1}_{i,t}
            w(u_i(t)),
            \eta
        \right\}
    },
    \label{eq:overlap_add}
\end{equation}
where $w(\cdot)$ is a nonnegative window function and $\eta>0$
prevents division by zero. Where the total window weight exceeds the numerical floor, normalization yields a weighted average of the contributing samples and avoids amplitude scaling from an unnormalized overlap sum.

For overlapping configurations without an explicit tapering window,
the contributing samples are combined through normalized overlap
averaging. Triangular and Hann configurations instead assign
position-dependent weights within each segment.

\label{sec:diagnostics}

Segmented generation introduces a boundary-junction risk because
separately generated windows need not agree where consecutive segments
are joined. We measure this local assembly artifact using
\begin{equation}
    J_{\mathrm{jump}}
    =
    \frac{1}{|\mathcal{J}|}
    \sum_{t_j\in\mathcal{J}}
    \left\|
        \tilde{S}(t_j)-\tilde{S}(t_j-1)
    \right\|_2,
    \label{eq:j_jump}
\end{equation}
where $\mathcal{J}$ denotes the set of assembly-junction indices. For a univariate series, the norm reduces to the absolute
difference between adjacent values. Unlike a global smoothness
measure, $J_{\mathrm{jump}}$ specifically targets discontinuities at
assembly junctions.

\paragraph{Proposition 1: Conditional boundary-junction reduction}
Consider two windows $A$ and $B$ blended in an overlap region as
$Y_r=(1-\lambda_r)A_{k+r}+\lambda_rB_r$, with $0\leq\lambda_r\leq1$.
Writing $\Delta A_r=A_{k+r+1}-A_{k+r}$,
$\Delta B_r=B_{r+1}-B_r$, and
$\Delta\lambda_r=\lambda_{r+1}-\lambda_r$, the local transition satisfies
\begin{equation}
\|Y_{r+1}-Y_r\|_2\leq
(1-\lambda_{r+1})\|\Delta A_r\|_2+
\lambda_{r+1}\|\Delta B_r\|_2+
|\Delta\lambda_r|\,\|B_r-A_{k+r}\|_2.
\label{eq:transition_bound}
\end{equation}
Thus, if within-window increments are bounded by $a$, cross-window
mismatch by $b$, and weight increments by $c$, the blended transition is
bounded by $a+bc$. It is smaller than a hard-concatenation jump of size
$H$ whenever $a+bc<H$.

This bound identifies the mechanism behind boundary reduction: gradual
weight changes limit the contribution of cross-window mismatch, while
within-window variation remains part of the assembled signal. Within the
two-window overlap, the bound therefore explains how gradual weighting can
moderate boundary transitions; $J_{\mathrm{jump}}$ aggregates these transitions
at the junctions defined by the assembly configuration.
The derivation is provided in \hyperref[app:diagnostic_rationale]{Appendix~C.1}.

\subsection{Boundary Continuity and Temporal Utility}
\label{sec:principle_utility}
\paragraph{Proposition 2: Boundary continuity does not determine temporal fidelity}
Let $\mathcal{B}=\bigcup_{t_j\in\mathcal{J}}\{t_j-1,t_j\}$ contain the
samples used by the junction diagnostic. For any perturbation $h$ with
$h_t=0$ on $\mathcal{B}$, the sequences $X$ and $X+h$ have identical
$J_{\mathrm{jump}}$. When unmeasured positions remain, perturbations at
those positions can change marginal and temporal statistics while
preserving the boundary score.

The equality follows directly from the definition of $J_{\mathrm{jump}}$.
A nonzero perturbation away from $\mathcal{B}$ can alter the value
distribution, adjacent increments, and lagged products used to compute
autocorrelation. Boundary continuity therefore leaves sequence properties
relevant to forecasting unspecified. The resulting TSTR response depends on the forecaster and the target task; we examine that response through the TSTR evaluations described below.
\hyperref[app:diagnostic_rationale]{Appendix~C.1} gives the supporting construction.

\subsection{Joint Evaluation Design}
\label{sec:joint_evaluation}
Boundary continuity captures only one aspect of an assembled sequence. We
therefore evaluate marginal distributional agreement and temporal fidelity
alongside the junction diagnostic. Let $\hat{X}\in\mathbb{R}^{T\times d}$
denote an assembled synthetic sequence and let $X$ denote the corresponding real
reference sequence. Marginal distributional discrepancy is summarized
using the Wasserstein distance.

Temporal dependence is assessed at a 24-hour physical lag. Let
$\ell_{24h}$ denote its sample count at the dataset resolution.
We then compute
\begin{equation}
    \Delta\mathrm{ACF}_{24h}
    =
    \left|
        \mathrm{ACF}_{\ell_{24h}}(\hat{X})
        -
        \mathrm{ACF}_{\ell_{24h}}(X)
    \right|.
    \label{eq:acf_deviation}
\end{equation}
These diagnostics deliberately capture different properties. A lower $J_{\mathrm{jump}}$ indicates lower measured boundary discontinuity at
junctions, whereas lower distributional or autocorrelation discrepancy
indicates closer agreement under the corresponding statistical
diagnostic. These diagnostics are interpreted as complementary descriptors of
assembly quality and are evaluated alongside downstream forecasting
error.

\label{sec:tstr_trtr}

Downstream utility is evaluated primarily using the
Train-on-Synthetic-Test-on-Real (TSTR) protocol. For an assembly
configuration $r$, a forecasting model is trained on the corresponding
assembled synthetic training sequence and evaluated on held-out
real observations. We report mean absolute error (MAE), with lower
values indicating lower forecasting error.

TSTR comparisons within an assembly grid reveal how assembly changes
downstream utility while the synthesis pipeline is otherwise held fixed. Matched Train-on-Real-Test-on-Real (TRTR) references provide the
corresponding real-training context for interpreting relative differences
among synthetic assembly settings.

A matched TRTR reference uses the same downstream forecasting task,
evaluation target, error scale, and test protocol as the corresponding
TSTR comparison, while replacing synthetic training data with the
matched real training data. TRTR provides the corresponding real-training reference for absolute utility and places TSTR performance on the same error scale.

To summarize assembly-induced variation for each dataset and
forecaster, we report the TSTR range over the evaluated grid:
\begin{equation}
    \mathcal{I}_{\mathrm{TSTR}}
    =
    \left[
        \min_{r\in\mathcal{R}}\mathrm{MAE}_{\mathrm{TSTR}}(r),
        \max_{r\in\mathcal{R}}\mathrm{MAE}_{\mathrm{TSTR}}(r)
    \right].
    \label{eq:tstr_range}
\end{equation}
Here $\mathcal{R}$ denotes the evaluated set of assembly configurations,
including naive concatenation. Each $\mathrm{MAE}_{\mathrm{TSTR}}(r)$
is the configuration-level mean under the aggregation protocol used for
the reported results.
The interval summarizes assembly-dependent variation in forecasting
performance, with matched TRTR providing the corresponding real-training
reference.

The lowest-MAE configurations in the grid identify candidate settings for
the target forecasting task. Together, the two propositions motivate the
three-part evaluation used throughout the experiments: boundary continuity
characterizes local junction behavior, fidelity diagnostics characterize
broader sequence properties, and TSTR measures task utility. Matched TRTR places that utility on the real-training error scale. The public-data regularities are subsequently examined in the operational dataset, while the second generator provides a separate cross-generator check of the forecaster-dependent assembly response.

\section{Experimental Design and Results}
\label{sec:results}
The experiments use the public datasets ETTh1, ETTm1, Weather, and Appliances to identify forecaster-dependent sequence-assembly regularities through joint evaluation of boundary continuity, statistical and temporal fidelity, and downstream TSTR utility. We then examine whether these regularities persist in an operational electricity-load forecasting task and assess the key forecaster-dependent response under a second generator architecture.

\subsection{Experimental Protocol}
\label{sec:experimental_setup}

\subsubsection{Datasets}
\label{sec:datasets}

The public-data analysis uses ETTh1, ETTm1, Weather, and Appliances to identify recurring assembly patterns across datasets and forecasters. These datasets cover electricity-transformer, environmental, and residential-energy time series at hourly, 15-minute, and 10-minute resolutions. An operational electricity-load dataset is subsequently used for domain-specific application and validation. All datasets use the target and preprocessing protocol associated with the reported experiments.

\begin{table}[htbp]
\centering
\caption{Summary of datasets used in the experimental evaluation.}
\label{tab:dataset_summary}
\small
\begin{adjustbox}{max width=\linewidth}
\begin{tabular}{lllll}
\toprule
Dataset & Domain & Resolution & Target & Role in study \\
\midrule
ETTh1 & Electricity transformer & Hourly & OT & Public-dataset regularity discovery \\
ETTm1 & Electricity transformer & 15 minutes & OT & Public-dataset regularity discovery \\
Weather & Environmental & 10 minutes & OT & Public-dataset regularity discovery \\
Appliances & Residential energy & 10 minutes & Appliances energy use & Public-dataset regularity discovery \\
Operational load & Electricity load & Hourly & Aggregate load & Operational application and validation \\
\bottomrule
\end{tabular}
\end{adjustbox}
\end{table}

\subsubsection{Sequence Assembly Settings}
\label{sec:assembly_settings}

For the four public datasets, the primary generator is a DP-SGD-trained
conditional 1D U-Net diffusion model that produces fixed-length synthetic
windows for subsequent sequence assembly. We evaluate assembly operators
$A_{\rho,w}$, where $\rho$ is the overlap rate and $w$ denotes the
window-weighting schemes. The candidate settings include naive concatenation,
overlap without tapering, triangular weighting, and Hann weighting.
Generator checkpoints are held fixed within each seed across the corresponding
assembly comparisons. The four public-dataset protocols use seven-day generated
windows and 70-day synthetic training sequences. The corresponding
$(L,T_{\mathrm{final}})$ settings are $(168,1680)$ for ETTh1,
$(672,6720)$ for ETTm1, and $(1008,10080)$ for both Weather and Appliances;
operational load uses its dataset-specific protocol.

For each stride, the minimum number of generated windows is used to
cover the target training length, and excess samples are deterministically
cropped. Calibration, where used, precedes overlap-add assembly;
its release protocol and privacy accounting are unchanged and described
in \hyperref[app:dp_details]{Appendix~B}. Public-dataset TSTR
assembly grids are provided in \hyperref[app:full_results]{Appendix~D},
while matched TRTR references for all five forecasters are provided
in \hyperref[app:complete_trtr]{Appendix~C};
the operational grid is provided in
\hyperref[app:public_operational_results]{Appendix~E}.
The generated-window count is
\begin{equation}
    M=\left\lceil\frac{T_{\mathrm{final}}-L}{k}\right\rceil+1,
    \label{eq:generated_window_count}
\end{equation}
with deterministic cropping to $T_{\mathrm{final}}$ when needed.

\subsubsection{Forecasting Models and Evaluation Metrics}
\label{sec:forecast_protocol}
\label{sec:metrics}

The four public evaluation datasets and the operational-load case use
LSTM, RNN, LightGBM, ARIMA, and Prophet. The detailed main-text endpoint
and TRTR comparisons highlight LSTM and ARIMA, while
Section~\ref{sec:forecaster_preferences} summarizes assembly
principles across all five forecasters. Five-forecaster TSTR
assembly grids for the public datasets are reported in
\hyperref[app:full_results]{Appendix~D},
while matched TRTR references for all five forecasters are
reported in \hyperref[app:complete_trtr]{Appendix~C}. The operational assembly grid is reported in
\hyperref[app:public_operational_results]{Appendix~E}, while the matched
operational TRTR references are summarized in Section~\ref{sec:operational_results}. TSTR
trains each forecaster on assembled synthetic sequences and evaluates
it on held-out real observations. Matched TRTR references provide the
real-training reference for the evaluated public datasets and the
operational comparison. We summarize boundary continuity
with $J_{\mathrm{jump}}$, fidelity with Wasserstein distance and the
dataset-appropriate autocorrelation deviation, and forecasting utility
with MAE. Three independently trained generator seeds and 20 independently sampled
assembled-sequence replicas per seed are used in the diffusion comparisons.
Results are averaged first within generator seed over the 20 synthetic-sequence
replicas and then across the three independently trained generator seeds.
Detailed forecasting-task construction, including model-specific recursive
or full-segment prediction procedures, is reported in
\hyperref[app:model_configs]{Appendix~A.6}.

\subsection{Sequence Assembly Effects on Boundary Continuity across Datasets}
\label{sec:boundary_reduction}

This experiment examines the boundary behavior motivated by Proposition 1 across the four public evaluation datasets.
\tabref{tab:boundary_reduction} summarizes the boundary-discontinuity reductions by comparing naive concatenation with the
lowest high-overlap $J_{\mathrm{jump}}$ value for each dataset.

\begin{table}[htbp]
\centering
\caption{Boundary discontinuity reduction across datasets. Values are
seed-averaged means; the assembly column reports the lowest
$J_{\mathrm{jump}}$ among the evaluated high-overlap settings.}
\label{tab:boundary_reduction}
\small
\begin{tabular}{lccc}
\toprule
Dataset & Naive $J_{\mathrm{jump}}$ & Evaluated assembly $J_{\mathrm{jump}}$ & Reduction \\
\midrule
ETTh1 & 4.7745 & 0.3964 & 91.7\% \\
ETTm1 & 4.1697 & 0.1412 & 96.6\% \\
Weather & 47.711 & 29.853 & 37.4\% \\
Appliances & 216.001 & 94.367 & 56.3\% \\
\bottomrule
\end{tabular}
\end{table}

Across the four datasets, the selected high-overlap configurations exhibit
lower measured boundary discontinuity than naive concatenation.

\subsection{Continuity--Fidelity--Utility Trade-offs across Datasets}
\label{sec:ett_tradeoff}
\label{sec:smoothness_fidelity}

\begin{figure}[!htbp]
    \centering
    \includegraphics[width=0.72\linewidth]{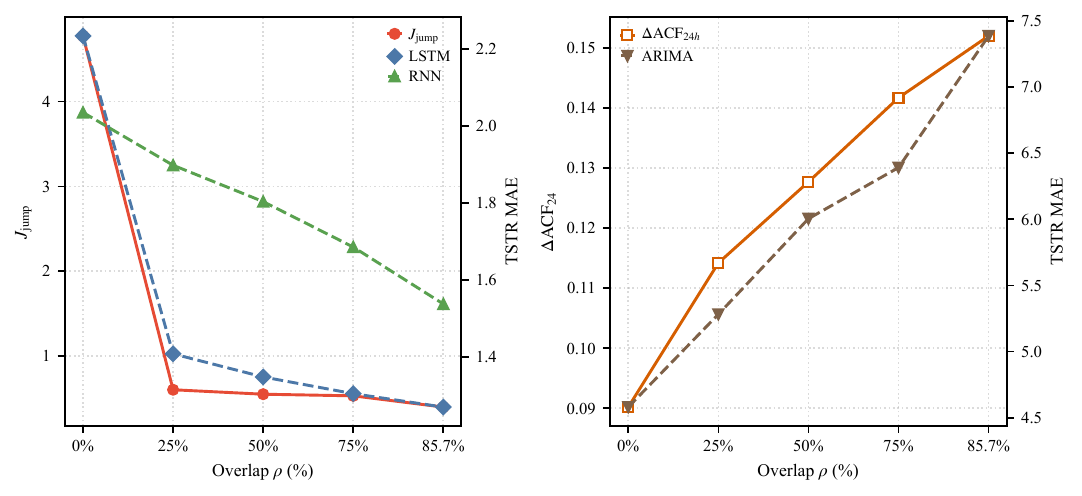}
    \caption{ETTh1 assembly path under triangular
    weighting. Boundary discontinuity decreases while the physical
    24-hour autocorrelation deviation changes; LSTM and ARIMA TSTR MAE
    show different downstream responses. The corresponding endpoint
    comparisons across the four public evaluation datasets are reported
    in \tabref{tab:multidataset_endpoints}.}
    \label{fig:smoothness_fidelity}
\end{figure}

The four public evaluation datasets are compared under a common
continuity--fidelity--utility protocol. \figref{fig:smoothness_fidelity}
uses ETTh1 to visualize the increasing-overlap
triangular assembly path, while \tabref{tab:multidataset_endpoints}
summarizes the corresponding endpoint comparisons across ETTh1, ETTm1,
Weather, and Appliances. The physical 24-hour autocorrelation lag is
represented by 24 samples for ETTh1, 96 samples for ETTm1, and 144 samples
for Weather and Appliances.

\tabref{tab:multidataset_endpoints} compares naive concatenation with
the high-overlap triangular candidate for all four public evaluation
datasets. Lower values indicate lower measured error or deviation. The
$\Delta\mathrm{ACF}_{24h}$ column denotes the dataset-specific
physical 24-hour lag described above.

\begin{table}[!htbp]
\centering
\caption{Continuity, fidelity, and forecasting utility for the endpoint
comparison across the four public evaluation datasets. The
$\Delta\mathrm{ACF}_{24h}$ column uses the physical 24-hour lag: 24 samples
for ETTh1, 96 for ETTm1, and 144 for Weather and Appliances. All values
are seed-averaged means; lower values indicate lower measured error or
deviation.}
\label{tab:multidataset_endpoints}
\small

\begin{adjustbox}{max width=\linewidth}
\begin{tabular}{llccccc}
\toprule
\multirow{2}{*}{Dataset}
& \multirow{2}{*}{Setting}
& \multicolumn{3}{c}{Diagnostics}
& \multicolumn{2}{c}{TSTR MAE} \\
\cmidrule(lr){3-5}
\cmidrule(lr){6-7}
&
& $J_{\mathrm{jump}}$
& Wasserstein
& $\Delta\mathrm{ACF}_{24h}$
& LSTM
& ARIMA \\
\midrule

ETTh1
& Naive
& 4.7745
& 1.7020
& 0.0902
& 2.175
& 4.578 \\

ETTh1
& High-overlap Triangular
& 0.3964
& 1.6805
& 0.1520
& 1.270
& 7.386 \\

ETTm1
& Naive
& 4.1697
& 1.6681
& 0.0363
& 3.998
& 1.513 \\

ETTm1
& High-overlap Triangular
& 0.1412
& 1.0624
& 0.0986
& 1.641
& 3.242 \\

Weather
& Naive
& 47.711
& 56.830
& 0.2312
& 11.518
& 15.767 \\

Weather
& High-overlap Triangular
& 29.853
& 56.803
& 0.3378
& 8.752
& 27.282 \\

Appliances
& Naive
& 216.001
& 29.729
& 0.0755
& 48.600
& 50.090 \\

Appliances
& High-overlap Triangular
& 94.367
& 33.598
& 0.0692
& 38.630
& 60.970 \\

\bottomrule
\end{tabular}
\end{adjustbox}
\end{table}

Across the four datasets, the evaluated high-overlap candidates have lower
measured $J_{\mathrm{jump}}$ than naive concatenation, while fidelity and
forecasting responses vary by dataset and forecaster. On ETTh1, ETTm1,
and Appliances, the high-overlap triangular candidate also has lower LSTM
MAE, whereas its ARIMA MAE is higher than the naive configuration. Weather
shows the same forecaster-dependent contrast in the endpoint comparison.
The complete assembly grids in \hyperref[app:full_results]{Appendix~D} further characterize the
dataset- and forecaster-dependent relationships among continuity, fidelity,
and forecasting utility. 

\FloatBarrier

\subsection{Forecasting Utility Relative to Matched Real-Data Training}
\label{sec:trtr_results}

\tabref{tab:trtr_main} places assembly-induced TSTR variation
alongside matched real-training references. For each public evaluation
dataset, it reports matched TRTR, naive TSTR, and the minimum--maximum
range of seed-averaged TSTR MAE for LSTM and ARIMA across the 13 evaluated
assembly settings. \hyperref[app:full_results]{Appendix~D} provides the
five-forecaster TSTR assembly grids, while
\hyperref[app:complete_trtr]{Appendix~C} provides the corresponding matched
TRTR references. Generator training is fixed within each dataset and seed,
so the range summarizes forecasting sensitivity to sequence assembly.

\begin{table}[htbp]
\centering
\caption{Matched TRTR references and assembly-induced TSTR ranges on
the four public evaluation datasets. All values are MAE. The table reports
LSTM and ARIMA; matched TRTR references for all five forecasters are provided
in \hyperref[app:complete_trtr]{Appendix~C}. The TSTR range spans the minimum
and maximum mean MAE across all 13 assembly settings, including naive
concatenation.}
\label{tab:trtr_main}
\small
\begin{adjustbox}{max width=\linewidth}
\begin{tabular}{llccc}
\toprule
Dataset & Forecaster & Matched TRTR & Naive TSTR &
TSTR range across assembly \\
\midrule
ETTh1 & LSTM & 1.291 & 2.175 & 1.270--2.175 \\
      & ARIMA & 0.702 & 4.578 & 4.578--7.770 \\
\midrule
ETTm1 & LSTM & 0.768 & 3.998 & 1.641--3.998 \\
      & ARIMA & 0.704 & 1.513 & 1.513--3.372 \\
\midrule
Weather & LSTM & 4.199 & 11.518 & 8.244--13.085 \\
        & ARIMA & 3.460 & 15.767 & 15.767--27.972 \\
\midrule
Appliances & LSTM & 27.760 & 48.600 & 38.630--48.600 \\
           & ARIMA & 23.490 & 50.090 & 50.090--71.100 \\
\bottomrule
\end{tabular}
\end{adjustbox}
\end{table}

The ranges show model- and dataset-dependent assembly sensitivity.
For example, ETTh1 LSTM spans 1.270--2.175 relative to a matched TRTR
reference of 1.291, whereas ETTh1 ARIMA spans 4.578--7.770 relative to a
TRTR reference of 0.702. The corresponding ETTm1, Weather, and Appliances
results likewise place assembly-induced TSTR variation on the matched
real-training error scale. Matched TRTR references for all five forecasters
are reported in \hyperref[app:complete_trtr]{Appendix~C}.

\FloatBarrier

\subsection{Forecaster-Dependent Assembly Principles}
\label{sec:forecaster_preferences}

\begin{figure}[!htbp]
    \centering
    % Remove the obsolete in-image title; the LaTeX caption supplies the title.
    \includegraphics[width=0.92\linewidth,trim=0 0 0 18,clip]{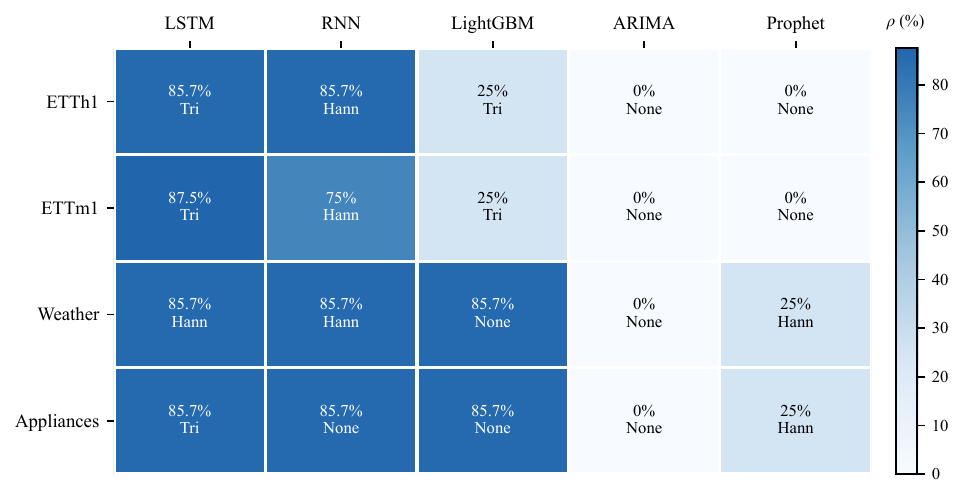}
    \caption{Minimum-MAE assembly configurations identified from the
    seed-averaged grids for the four public evaluation datasets. Each cell
    represents one dataset--forecaster pair. Color intensity represents
    overlap rate $\rho$; text labels denote Tri (Triangular weighting),
    Hann (Hann weighting), and None (no explicit windowing).}
    \label{fig:assembly_preferences}
\end{figure}

\figref{fig:assembly_preferences} summarizes the
minimum-MAE assembly configuration for each dataset--forecaster pair across
the four public datasets. Three forecaster-dependent regularities emerge.
First, LSTM and RNN generally favor intermediate-to-high overlap. Second,
ARIMA attains its lowest reported MAE at zero overlap on all four public
datasets, while Prophet similarly favors zero or low overlap. Third, LightGBM
shows stronger dataset dependence across overlap and weighting choices.

These regularities provide a forecaster-aware basis for prioritizing assembly candidates. Intermediate-to-high overlap provides a natural starting region for recurrent models, whereas zero-to-low overlap provides a natural starting region for ARIMA and Prophet; LightGBM requires broader task-specific evaluation. 

\FloatBarrier

\subsection{Operational Electricity-Load Application}
\label{sec:operational_results}

We apply the public-dataset assembly principles to aggregate
electricity-load forecasting over a 24-hour real evaluation window using
assembled synthetic hourly sequences. Short-term load forecasts can inform
generation dispatch, electricity procurement, and demand-side load control.
\tabref{tab:operational_trtr} reports matched TRTR MAE, naive TSTR MAE, and
the TSTR range over all evaluated assembly settings. The full operational grid
is provided in \hyperref[app:public_operational_results]{Appendix~E}.

\begin{table}[htbp]
\centering
\caption{Operational electricity-load forecasting results. All entries
are MAE; the TSTR range spans the evaluated assembly settings.}
\label{tab:operational_trtr}
\small
\begin{tabular}{lccc}
\toprule
Forecaster & TRTR MAE & Naive TSTR MAE & TSTR range \\
\midrule
LSTM & 0.955 & 1.685 & 1.147--1.724 \\
RNN & 1.306 & 1.961 & 1.640--2.645 \\
LightGBM & 0.882 & 0.900 & 0.900--1.435 \\
ARIMA & 1.542 & 2.667 & 2.599--3.092 \\
Prophet & 1.399 & 2.220 & 2.220--3.148 \\
\bottomrule
\end{tabular}
\end{table}
\FloatBarrier
The operational results exhibit the same broad forecaster-dependent tendency identified across the public datasets. LSTM and RNN achieve their lowest MAE at 75\% and 85.7\% overlap, respectively, consistent with the higher-overlap preference of recurrent forecasters. ARIMA and Prophet attain their minima at low or zero overlap, consistent with their public-dataset tendency, whereas LightGBM again shows task-specific behavior. These results provide domain-specific support for the forecaster-dependent assembly regularities and illustrate how the public-dataset findings can guide candidate prioritization for downstream evaluation.

\subsection{Cross-generator Check with DP-ConvVAE}
\label{sec:vae_results}

The DP-ConvVAE comparison examines whether the forecaster-dependent assembly response persists under a second
generator architecture using ETTh1 and ETTm1. Three settings compare naive
concatenation ($C0$), high-overlap equal weighting ($C1$), and Triangular
weighting at the same overlap ($C2$); full metric results are reported in
\hyperref[app:generator_comparison]{Appendix~F}.

\begin{table}[!htbp]
\centering
\caption{DP-ConvVAE results across three assembly settings. Values are means; entries are ordered C0 $\rightarrow$ C1 $\rightarrow$ C2.}
\resizebox{\linewidth}{!}{%
\begin{tabular}{lccc}
\toprule
Dataset & $J_{\mathrm{jump}}$ & LSTM MAE & ARIMA MAE \\
\midrule
ETTh1 & 0.6892 $\rightarrow$ 0.1233 $\rightarrow$ 0.0549 & 3.080 $\rightarrow$ 2.947 $\rightarrow$ 2.814 & 2.952 $\rightarrow$ 3.215 $\rightarrow$ 4.154 \\
ETTm1 & 0.4812 $\rightarrow$ 0.2125 $\rightarrow$ 0.1770 & 2.848 $\rightarrow$ 2.822 $\rightarrow$ 2.776 & 4.020 $\rightarrow$ 4.225 $\rightarrow$ 4.319 \\
\bottomrule
\end{tabular}%
}
\end{table}

Across both datasets, lower $J_{\mathrm{jump}}$ across the three assembly settings is accompanied by lower LSTM MAE and higher ARIMA MAE. This reproduces the contrasting forecaster responses to sequence assembly under a second generator architecture.

\section{Discussion and Implications for Forecasting Systems}
\label{sec:discussion}

The experiments reveal a consistent system-level picture of post-generation
sequence assembly. Overlap and window weighting strongly influence local
junction continuity, but the same assembly change can affect temporal fidelity
and TSTR utility differently. Most importantly, the downstream response is
forecaster dependent. This dependence turns sequence assembly from a generic
preprocessing choice into a forecaster-aware model-selection problem.

\subsection{Why Local Smoothness Is Insufficient for Assembly Selection}
\label{sec:discussion_smoothness}

Proposition 1 explains why overlap and gradual weighting can reduce local
junction transitions, and the public-dataset experiments confirm large
reductions in $J_{\mathrm{jump}}$. Proposition 2 complements this mechanism by
showing that a junction score leaves broader sequence properties unresolved.
The empirical results connect these two observations: lower boundary
discontinuity can coincide with either lower or higher forecasting error,
depending on the forecaster, and individual fidelity diagnostics show the same
lack of one-to-one correspondence with TSTR utility. Continuity and fidelity
therefore function as complementary descriptors, while downstream utility
remains the decisive criterion for assembly selection.

\subsection{Forecaster-Dependent Assembly as a Model-Selection Problem}
\label{sec:discussion_forecaster}

The complete five-forecaster grids expose structure that is hidden by any single forecasting model. Recurrent forecasters generally favor intermediate-to-high overlap, where averaging and tapering can moderate local junction effects while preserving useful temporal patterns. ARIMA and Prophet favor zero or low overlap more often, indicating greater sensitivity to the temporal changes introduced by aggressive blending. LightGBM shows stronger dataset dependence across assembly settings. This behavior is consistent with its lag-feature-based prediction structure: rather than explicitly maintaining a recurrent temporal state, LightGBM operates on a fixed set of lagged observations and calendar features. Consequently, improvements in local junction continuity need not translate directly into improved predictive features, and its preferred assembly configuration can depend more strongly on the dataset-specific lag structure. These patterns provide a practical ordering of assembly configurations before target-task validation rather than a single globally optimal assembly rule.

\subsection{From Public-Dataset Regularities to Operational Validation}
\label{sec:engineering_guidance}

The public datasets provide the discovery stage for forecaster-dependent assembly regularities, while the operational electricity-load experiment provides a domain-specific validation stage. The same broad model-specific tendencies persist: recurrent models favor higher overlap, ARIMA and Prophet favor lower overlap, and LightGBM remains more task dependent. This consistency shows how public-dataset regularities can inform the prioritization of assembly candidates in a domain-specific forecasting system.

In practice, these regularities can be used to prioritize a smaller set of assembly configurations for target-task evaluation with the intended forecaster. Downstream TSTR utility provides the basis for configuration selection, while continuity and fidelity diagnostics offer complementary interpretation and matched TRTR provides the corresponding real-training reference.

\section{Limitations}
\label{sec:limitations}

The experiments use a controlled assembly space defined by the evaluated
overlap rates and window-weighting schemes, with generated-window length and
final assembled training length fixed within each dataset protocol. Extending
the design space to adaptive weighting, learned assembly operators, and
alternative window lengths would broaden the range of sequence-construction
behaviors covered by the framework.

The operational electricity-load application uses a 24-hour real evaluation
window. Evaluation across multiple rolling windows and longer operational periods would further characterize the temporal stability of the forecaster-dependent assembly patterns under changing load conditions.

\section{Conclusion}

This study formulated post-generation sequence assembly as an explicit design
stage in forecasting pipelines based on differentially private synthetic time
series. By holding the trained generator fixed and varying overlap rates and window-weighting schemes, we isolated how assembly changes boundary continuity,
statistical and temporal fidelity, and downstream TSTR forecasting utility.

Across ETTh1, ETTm1, Weather, and Appliances, increased overlap generally
reduced junction discontinuity, but improvements in continuity or individual
fidelity diagnostics did not consistently reduce forecasting error. Complete
five-forecaster assembly grids revealed a clear forecaster-dependent pattern:
recurrent forecasters more often favored intermediate-to-high overlap, ARIMA
and Prophet favored zero or low overlap, and LightGBM showed stronger dataset
dependence. Matched TRTR references placed the resulting TSTR variation on the
corresponding real-training error scale.

The public-dataset regularities were further examined in an operational electricity-load forecasting task, where the same broad forecaster-dependent assembly tendencies persisted. A DP-ConvVAE comparison further reproduced a forecaster-dependent assembly response under a second generator architecture. Together, these results establish sequence assembly as a forecaster-aware model-selection problem and show how multi-dataset regularities can inform domain-specific deployment of differentially private synthetic time series.

\newpage
\appendix

\counterwithin{table}{section}
\counterwithin{figure}{section}

\section*{Appendix Roadmap}
\phantomsection
\label{app:roadmap}
\hyperref[app:implementation]{Appendix~A} reports reproducibility and
implementation details for the four main public datasets and downstream
forecasting protocols;
\hyperref[app:dp_details]{Appendix~B} reports differential-privacy and
calibration details;
\hyperref[app:supporting_analysis]{Appendix~C} provides supporting
methodological analyses and complete matched TRTR references;
\hyperref[app:full_results]{Appendix~D} reports the complete five-forecaster
TSTR assembly grids for ETTh1, ETTm1, Weather, and Appliances;
\hyperref[app:public_operational_results]{Appendix~E} reports the operational
electricity-load application; and
\hyperref[app:generator_comparison]{Appendix~F} reports the DP-ConvVAE cross-generator check.

\FloatBarrier

\section{Reproducibility and Implementation Details}
\label{app:implementation}

\subsection{Main Public Datasets and Chronological Splits}
\label{app:main_public_datasets}

The main public evaluation uses four equal-status datasets: ETTh1, ETTm1,
Weather, and Appliances. ETTh1 is hourly and ETTm1 is recorded at 15-minute
resolution; Weather and Appliances are recorded at 10-minute resolution.
Dataset-specific sources, targets, and settings are detailed in
\hyperref[app:dataset_settings]{Appendix~A.3}.

\paragraph{Training-period normalization}
The target series is normalized using a MinMaxScaler fitted exclusively on
the generator-training interval. Let $x_{\min,\mathrm{train}}$ and
$x_{\max,\mathrm{train}}$ denote the minimum and maximum estimated from
this interval. The transformation is
\begin{equation}
    x'_t=
    \frac{x_t-x_{\min,\mathrm{train}}}
    {x_{\max,\mathrm{train}}-x_{\min,\mathrm{train}}}.
    \label{eq:training_minmax}
\end{equation}
The held-out evaluation period is excluded from scaler fitting, preserving
the separation between training and evaluation data. The fitted minimum
and maximum are data-dependent preprocessing quantities and are not
separately privatized. Accordingly, the formal DP-SGD guarantee in
Section~\ref{sec:privacy_scope} is defined on the normalized extracted
window dataset supplied to the training mechanism; MinMax parameter
estimation lies outside that guarantee. Calibration statistics are
estimated from clipped original target observations using the separate
mechanism in \hyperref[app:dp_details]{Appendix~B}.

\paragraph{Generator-training window construction}
Generator training uses fixed-length windows extracted from
$\mathcal{D}_{\mathrm{norm}}$ with $s_{\mathrm{train}}=1$.
Consequently, for a training sequence of length $T_{\mathrm{train}}$ and
window length $L$, the number of training records is
\begin{equation}
    N=T_{\mathrm{train}}-L+1.
    \label{eq:training_record_count}
\end{equation}
These overlapping windows are treated as separate training records by the
implemented DP-SGD accountant, so the corresponding guarantee is defined
on this normalized window dataset. The extraction stride $s_{\mathrm{train}}$ is
distinct from the placement stride $k$ used after generation.

\begin{table}[htbp]
\centering
\caption{Chronological data intervals for the four main public datasets.}
\label{tab:chronological_intervals}

\footnotesize
\setlength{\tabcolsep}{5pt}
\renewcommand{\arraystretch}{1.12}

\begin{adjustbox}{max width=\linewidth}
\begin{tabular}{@{}lll@{}}
\toprule
Dataset (sampling) & Training source interval & Real test interval \\
\midrule
ETTh1 (1 h) &
2017-07-01 00:00--2017-09-08 23:00 &
2017-09-09 00:00--2017-09-09 23:00 \\

ETTm1 (15 min) &
2017-07-01 00:00--2017-09-08 23:45 &
2017-09-09 00:00--2017-09-09 23:45 \\

Weather (10 min) &
2020-07-01 00:00--2020-09-08 23:50 &
2020-09-09 00:00--2020-09-09 23:50 \\

Appliances (10 min) &
2016-01-11 17:00--2016-03-21 16:50 &
2016-03-21 17:00--2016-03-22 16:50 \\

\bottomrule
\end{tabular}
\end{adjustbox}

\vspace{2pt}

\raggedright\footnotesize
\textit{Note:} Scaler fitting uses the training source interval; the real test
interval immediately follows and is excluded from generator training and scaler fitting.

\end{table}

\subsection{Sequence Assembly Protocols for the Main Public Datasets}
\label{app:assembly_protocols}

The public-dataset protocols are matched by physical duration: generated
windows span seven days and assembled downstream training sequences
span 70 days. Sample counts therefore scale with sampling resolution.
The corresponding $(L,T_{\mathrm{final}})$ settings are summarized in
\tabref{tab:assembly_length_protocol}. For each stride, $M$ is the
minimum generated-window count needed to cover $T_{\mathrm{final}}$ according
to Eq.~\ref{eq:generated_window_count}, and
$T_{\mathrm{raw}}=L+(M-1)k$ is the pre-crop length. The general crop rule
retains the first $T_{\mathrm{final}}$ samples whenever a configuration
exceeds that length.

\begin{table}[htbp]
    \centering
    \caption{Sequence assembly protocols for the four main public datasets.
    Here $k$ is the placement stride, $\rho$ is the overlap rate,
    $M$ is the minimum generated-window count, and $T_{\mathrm{final}}$
    is the assembled downstream training length.}
    \label{tab:assembly_length_protocol}
    \scriptsize
    \begin{adjustbox}{max width=\linewidth}
    \begin{tabular}{llccccc}
        \toprule
        Dataset & Sampling & $L$ & Placement strides $k$ & $\rho$ &
        Generated-window counts $M$ & $T_{\mathrm{final}}$ \\
        \midrule
        ETTh1 & 1 h & 168 & $\{168,126,84,42,24\}$ & $\{0\%,25\%,50\%,75\%,85.7\%\}$ & $\{10,13,19,37,64\}$ & 1680 \\
        ETTm1 & 15 min & 672 & $\{672,504,336,168,84\}$ & $\{0\%,25\%,50\%,75\%,87.5\%\}$ & $\{10,13,19,37,73\}$ & 6720 \\
        Weather & 10 min & 1008 & $\{1008,756,504,252,144\}$ & $\{0\%,25\%,50\%,75\%,85.7\%\}$ & $\{10,13,19,37,64\}$ & 10080 \\
        Appliances & 10 min & 1008 & $\{1008,756,504,252,144\}$ & $\{0\%,25\%,50\%,75\%,85.7\%\}$ & $\{10,13,19,37,64\}$ & 10080 \\
        \bottomrule
    \end{tabular}
    \end{adjustbox}
\end{table}

\subsection{Dataset-Specific Settings}
\label{app:dataset_settings}

\paragraph{ETT benchmarks}
The ETT datasets are public benchmarks introduced by
\citet{zhou2021informer}.
The OT target is evaluated at hourly resolution for ETTh1 and 15-minute
resolution for ETTm1.

\paragraph{Weather}
The Weather series follows the Autoformer forecasting benchmark suite~\citep{wu2021autoformer}
and originates from measurements collected at the Max Planck Institute for
Biogeochemistry in Jena, Germany. It uses the OT target at 10-minute resolution;
the physical 24-hour diagnostic therefore corresponds to 144 samples. Within
each seed, the generator checkpoint is fixed across the assembly sweep.
The assembly design contains 13 valid configurations: the 0\%
None baseline and positive-overlap None, Triangular, and Hann settings.
The complete sweep is reported in \hyperref[app:weather_results]{Appendix~D.3}.

\paragraph{Appliances}
The Appliances experiment uses the UCI Appliances Energy Prediction dataset
introduced by \citet{candanedo2017appliances}. The target is
Appliances energy use at 10-minute resolution, with a real evaluation window
of 144 samples (24 hours), beginning at 2016-03-21 17:00:00. Each generator
seed supplies 20 sequence replicas evaluated on the same fixed real test
window. Dataset-specific DP-SGD and calibration parameters are reported in
\hyperref[app:dp_details]{Appendix~B}. The complete sweep is reported in
\hyperref[app:appliances_results]{Appendix~D.4}.

\subsection{Diffusion Architecture and DP-SGD Training}

For the ETT, Weather, and Appliances experiments, the primary generator is a
conditional 1D U-Net diffusion model trained on fixed-length univariate
target windows. The target is OT for ETT and Weather, and Appliances energy
use for Appliances.
The model uses a cosine diffusion schedule with $1000$ diffusion steps.
We use Group Normalization rather than Batch Normalization to avoid
interactions between batch statistics and per-example gradient clipping.
The conditional model uses categorical embeddings for granularity, user
type, hour, day of week, day, month, and holiday indicators, with the
dataset-specific target channel and calibration scale described above.

\subsection{Compute Resources}

Generator training experiments were run on a single NVIDIA A100 GPU. The diffusion generator was trained with DP-SGD with batch size 96 for 200 epochs, using the architecture and optimization settings reported in \tabref{tab:diffusion_hparams}. For each dataset and random seed, one generator training run was performed, and the resulting synthetic windows were reused across the assembly grid. Sequence assembly, diagnostic metric computation, and downstream forecasting evaluations were run after generator training and required substantially less compute than generator training.

A typical generator training run on one A100 GPU took approximately 10 hours for ETTh1 and 14 hours for ETTm1. All generator experiments use seeds 65, 66, and 67; each seed has one generator training run per dataset, and the resulting synthetic windows are reused across that dataset's assembly grid. Downstream LSTM and RNN training used the settings in \hyperref[app:model_configs]{Appendix~A.6}; LightGBM, ARIMA, and Prophet were run on CPU.

\begin{table}[htbp]
    \centering
    \caption{Diffusion architecture and DP-SGD training hyperparameters.}
    \label{tab:diffusion_hparams}
    \scriptsize
    \begin{tabular}{lc}
        \toprule
        Hyperparameter & Value \\
        \midrule
        Backbone & Conditional 1D U-Net \\
        Input channels & 1 \\
        Output channels & 1 \\
        Base model channels & 144 \\
        Channel multipliers & $1,2,3,4$ \\
        Number of residual blocks & 2 \\
        Attention resolutions & $2,3,4$ \\
        Diffusion steps & 1000 \\
        Noise schedule & Cosine \\
        $\beta_{\min}$, $\beta_{\max}$ & $10^{-4}$, $0.02$ \\
        Optimizer & Adam \\
        Learning rate & $5\times 10^{-5}$ \\
        Batch size & 96 \\
        Training-window extraction stride $s_{\mathrm{train}}$ & 1 \\
        Training epochs & 200 \\
        Outlier clipping quantile & 0.999 \\
        Hardware & A100 \\
        \bottomrule
    \end{tabular}
\end{table}

\FloatBarrier

\subsection{Downstream Model Configurations}
\label{app:model_configs}

Under TSTR, each downstream forecasting model is trained separately on
each assembled-sequence replica and evaluated on the same fixed
real test set. Each public-dataset assembled-sequence replica spans 70 days,
corresponding to 1680 samples for ETTh1, 6720 samples for ETTm1, and
10080 samples for Weather and Appliances. The real evaluation window spans
24 hours: 24 samples for hourly data, 96 samples for 15-minute data, and
144 samples for 10-minute data. For every assembly setting and
generator checkpoint, 20 sequence replicas are evaluated. All four
public datasets use generator seeds 65, 66, and 67. Section~\ref{sec:metrics}
distinguishes replica-level evaluation from generator-seed summaries.

\paragraph{Forecasting-task construction}
LSTM and RNN use the previous 168 observations as input for one-step-ahead
prediction. LightGBM likewise uses 168 lagged target values for its
one-step-ahead prediction. Because the datasets have different sampling resolutions, this fixed 168-observation input window corresponds to 168 h for ETTh1, 42 h for
ETTm1, and 28 h for both Weather and Appliances. This fixed
sample-count protocol is retained throughout the public-dataset
downstream evaluation. During testing, each predicted target value is appended to the
history and used to construct the next input window, so evaluation over the
real segment is recursive. Their inputs also include sine/cosine encodings
of hour and day of week, together with a weekend indicator. LightGBM likewise
uses the corresponding calendar features; each prediction enters the lag
history for the next step during recursive evaluation. ARIMA is fitted once to each
training sequence and calls \texttt{forecast(steps=test\_len)} to generate
the complete evaluation segment, without refitting at each test step. For ARIMA, the seasonal period follows the physical 24-hour cycle at the
corresponding sampling resolution: 24 samples for hourly data, 96 samples
for 15-minute data, and 144 samples for 10-minute data.
Prophet is fitted once to the complete training sequence and predicts the
timestamps of the entire real evaluation segment in one call; it does not
use an explicit fixed lag window. For every forecaster, MAE is the mean of
$|y_{\mathrm{pred}}-y_{\mathrm{true}}|$ over all points in the real
evaluation window. Under TSTR, MAE is first computed separately for each
sequence replica and then aggregated using the seed/replica hierarchy
described in Section~\ref{sec:metrics}.

\begin{table}[htbp]
    \centering
    \caption{Downstream forecasting model hyperparameters.}
    \label{tab:downstream_hparams}
    \small
    \setlength{\tabcolsep}{4pt}
    \begin{adjustbox}{max width=\linewidth}
\begin{tabular}{lp{7.2cm}cc}
        \toprule
        Model & Hyperparameters & Optimizer / Solver & Epochs \\
        \midrule
        LSTM & hidden size 64, 2 layers, dropout 0.1 & Adam & 80 \\
        RNN & hidden size 64, 1 layer, dropout 0.0 & Adam & 80 \\
        LightGBM & 31 leaves, max depth $-1$, 200 estimators, learning rate 0.05, subsample 0.9, column sample 0.9 & LightGBM & -- \\
        ARIMA & order $(2,1,2)$; seasonal order $(1,0,1,s)$, with
        $s=24$ for hourly data, $s=96$ for 15-minute data, and
        $s=144$ for 10-minute data
        & L-BFGS-B & -- \\
        Prophet & additive seasonality; changepoint prior scale 0.05 & Prophet default & -- \\
        \bottomrule
    \end{tabular}
\end{adjustbox}
\end{table}

\FloatBarrier

\section{Differential Privacy and Calibration Details}
\label{app:dp_details}

\subsection{DP-SGD Training}

The diffusion generator is trained using DP-SGD with per-example
gradient clipping, where each example is one normalized fixed-length window record,
and Gaussian noise injection. We use a clipping norm of $C=1.0$, a batch
size of 96, 200 training epochs, and
$\delta_{\mathrm{train}}=10^{-5}$, with target
$\epsilon_{\mathrm{train}}=9$. The DP-SGD noise multiplier is
calibrated separately for each dataset using the Opacus RDP accountant
to satisfy the target
$(\epsilon_{\mathrm{train}},\delta_{\mathrm{train}})$ guarantee under
the corresponding training schedule on $\mathcal{D}_{\mathrm{win}}$.
The data-dependent normalization preceding window extraction is described
in \hyperref[app:implementation]{Appendix~A} and is outside this accounting.
The resulting dataset-specific
noise multipliers are reported in
\tabref{tab:dp_noise_multipliers}.

\begin{table}[htbp]
    \centering
    \caption{Dataset-specific DP-SGD noise multipliers used for generator
    training. The values are calibrated using Opacus to satisfy the target
    $(\epsilon_{\mathrm{train}},\delta_{\mathrm{train}})$ guarantee under
    the corresponding training schedule.}
    \label{tab:dp_noise_multipliers}
    \small
    \begin{tabular}{lc}
        \toprule
        Dataset & Noise multiplier $\sigma$ \\
        \midrule
        ETTh1 & 2.019 \\
        ETTm1 & 2.078 \\
        Weather & 2.024 \\
        Appliances & 2.024 \\
        Operational load & 2.041 \\
        \bottomrule
    \end{tabular}
\end{table}

Privacy loss is tracked using the same RDP accounting framework. The
placement stride $k$, generated-window count $M$, and deterministic
downstream crop are applied only after generation and do not change the
DP-SGD training dataset.

\subsection{Calibration Releases}

This section provides the implementation details for the calibration
setting described in Section~\ref{sec:privacy_scope}. The calibration
code operates on the original target observations at fixed timestamps
under the fixed-grid observation adjacency used by its bounded queries.
A pair of neighboring inputs differs in one timestamped observation.
Clipping is applied before the calibration statistics are computed. The
raw clipping bound is dataset-specific: $B=40.0$ for the ETT OT runs,
$B=550.0$ for Weather, $B=1100.0$ for Appliances, and
$B=6\times10^{7}$ for the operational-load run.
These bounds were selected a priori from typical value ranges in comparable
public or external reference data and were fixed before applying the
calibration mechanism to the corresponding records. The calibration release
uses $\epsilon_{\mathrm{cal}}=1$ and
$\delta_{\mathrm{cal}}=10^{-6}$.
These parameters describe the calibration mechanism separately from
DP-SGD on normalized window records, consistent with
Section~\ref{sec:privacy_scope}.

The main calibration mode uses noisy per-hour mean estimates together with a noisy global second-moment estimate. Let $\mathcal{G}=\{0,\ldots,23\}$ denote the 24 hour-of-day groups. For each group $g\in\mathcal{G}$, we compute the clipped hourly sum and release a noisy sum using the Gaussian mechanism:
\[
    \widetilde{s}_{g}
    =
    \sum_{t:h_t=g}
    \mathrm{clip}(x_t;0,B)
    +
    \mathcal{N}(0,\tau_{\mu}^{2}).
\]
The noisy hourly mean is then
\[
    \bar{\mu}_{g}
    =
    \frac{\widetilde{s}_{g}}{\max\{n_g,1\}},
\]
where $n_g$ is the number of observations in hour group $g$. The
timestamp grid and hour assignments are treated as fixed public design
information, so $n_g$ is not privately released. The per-contribution
bound is $B$; the sensitivity used to calibrate the Gaussian mechanism
is the sensitivity of the complete hourly-sum query under the fixed-grid
observation relation. Division by the group count is post-processing.

For scale calibration, we release a noisy global second moment:
\[
    \bar{q}
    =
    \frac{
        \sum_t \mathrm{clip}(x_t;0,B)^2
        +
        \mathcal{N}(0,\tau_q^2)
    }{
        \max\{n,1\}
    }.
\]
The corresponding noisy global standard deviation is computed as
\[
    \bar{\sigma}
    =
    \sqrt{
        \max\{
            \bar{q}-\bar{\mu}_{\mathrm{global}}^2,
            \sigma_{\min}^{2}
        \}
    }.
\]
The global second-moment release uses the sensitivity of the complete
second-moment query under the same fixed-grid observation relation; $B^2$
is the per-contribution bound. The quantity
$\bar{\mu}_{\mathrm{global}}$ is derived from the already
noise-perturbed hourly mean statistics and the fixed group counts, so
its use in the scale calculation does not introduce an additional query
to the raw target values.

The calibration parameter is split between the hourly mean release and
the global second-moment release:
\[
    \epsilon_{\mu}=\alpha\epsilon_{\mathrm{cal}},
    \qquad
    \epsilon_q=(1-\alpha)\epsilon_{\mathrm{cal}},
\]
and
\[
    \delta_{\mu}=\frac{\delta_{\mathrm{cal}}}{2},
    \qquad
    \delta_q=\frac{\delta_{\mathrm{cal}}}{2}.
\]
In the implementation, $\alpha=0.7$ unless otherwise stated. Thus,
$\epsilon_{\mu}=0.7$ and $\epsilon_q=0.3$. The Gaussian noise scales
$\tau_{\mu}$ and $\tau_q$ are calibrated from the corresponding
fixed-grid observation sensitivities and these allocated calibration
parameters.

The released noisy hourly means are optionally smoothed using fixed shrinkage and circular moving-average parameters. This smoothing uses only noisy DP estimates and fixed hyperparameters, and is therefore post-processing. Calibration then applies a global scale adjustment using the noisy global standard deviation and a per-hour shift using the noisy smoothed hourly means.

\begin{table}[htbp]
    \centering
    \caption{Calibration release parameters used in the main sequence-assembly experiments.}
    \label{tab:calibration_params}
    \scriptsize
    \begin{adjustbox}{max width=\linewidth}
    \begin{tabular}{lp{6cm}}
        \toprule
        Parameter & Value \\
        \midrule
        Calibration mode & per-hour mean with global scale adjustment \\
        Number of hour groups & 24 \\
        Released group-wise statistics & noisy per-hour means \\
        Released global statistics & noisy global second moment \\
        Calibration parameter $\epsilon_{\mathrm{cal}}$ & 1.0 \\
        Calibration failure probability $\delta_{\mathrm{cal}}$ & $10^{-6}$ \\
        Mean-budget fraction $\alpha$ & 0.7 \\
        Second-moment budget fraction $1-\alpha$ & 0.3 \\
        Mean release budget & $\epsilon_{\mu}=0.7\epsilon_{\mathrm{cal}}$ \\
        Second-moment release budget & $\epsilon_q=0.3\epsilon_{\mathrm{cal}}$ \\
        Mean release failure probability & $\delta_{\mu}=\delta_{\mathrm{cal}}/2$ \\
        Second-moment release failure probability & $\delta_q=\delta_{\mathrm{cal}}/2$ \\
        Raw clipping bound $B$ & ETT: 40.0; Weather: 550.0; Appliances: 1100.0; operational load: $6\times10^{7}$ \\
        Calibration seed & 0 \\
        Smoothing parameter & $\kappa=5$ \\
        Smoothing window & 3 \\
        Smoothing privacy cost & post-processing of noisy released means \\
        Calibration mechanism & Gaussian mechanism \\
        \bottomrule
    \end{tabular}
\end{adjustbox}
\end{table}

The main experiments treat assembly hyperparameters as prespecified evaluation choices.

\FloatBarrier

\section{Supporting Methodological Analysis}
\label{app:supporting_analysis}

\subsection{Rationale for Diagnostic Propositions}
\label{app:diagnostic_rationale}

\paragraph{Proposition 1}
Using the notation in Section~\ref{sec:principle_boundary}, subtraction and
regrouping yield
\[
Y_{r+1}-Y_r=(1-\lambda_{r+1})\Delta A_r+
\lambda_{r+1}\Delta B_r+\Delta\lambda_r(B_r-A_{k+r}).
\]
The triangle inequality gives Eq.~\ref{eq:transition_bound}. With the
stated bounds, the first two terms sum to at most $a$ and the third is
at most $bc$. Comparing this bound with $H$ gives the sufficient condition
$a+bc<H$ for a smaller local transition.

\paragraph{Proposition 2}
For fixed $\mathcal{J}$, take $q\notin\mathcal{B}$ and set
$h=\gamma e_q$, where $e_q$ is the coordinate vector at $q$ and
$\gamma\ne0$. Every measured junction pair is unchanged, so
$J_{\mathrm{jump}}(X+h)=J_{\mathrm{jump}}(X)$. For example, starting with a
constant sequence, this perturbation changes its sample marginal
distribution and first differences while leaving its junction score zero.
For nonconstant sequences, perturbations can also change lagged covariance.
The construction establishes that the junction score does not determine
these sequence properties. Forecasting consequences are assessed by the
TSTR comparisons in Section~\ref{sec:results}.

\subsection{Additional Diagnostic Relationship Analysis}
\label{app:diagnostic_corr}

\begin{figure}[htbp]
    \centering
    \includegraphics[width=\linewidth]{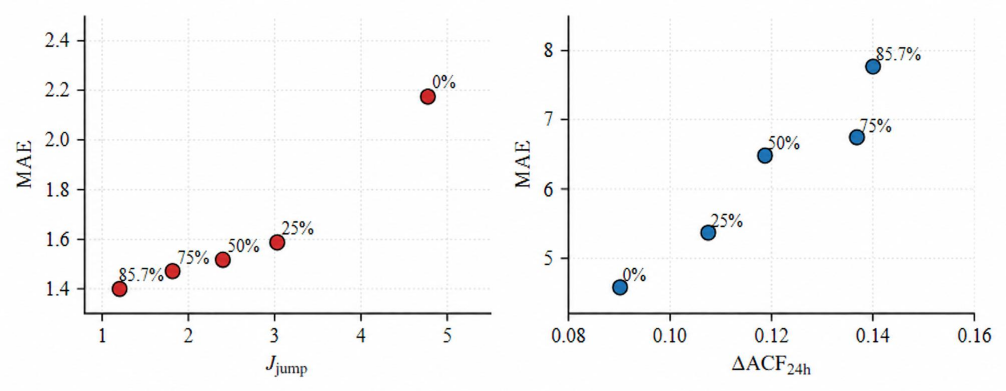}
    \caption{
    Additional diagnostic relationship analysis.
    Left: relationship between boundary discontinuity ($J_{\mathrm{jump}}$) and TSTR MAE for LSTM.
    Right: relationship between 24-hour autocorrelation deviation ($\Delta \mathrm{ACF}_{24h}$) and TSTR MAE for ARIMA.
    Each point corresponds to one overlap configuration in the evaluated assembly path.
    }
    \label{fig:mechanism_analysis_app}
\end{figure}

\figref{fig:mechanism_analysis_app} examines associations
between assembly diagnostics and downstream forecasting errors along
the ETTh1 triangular-window assembly path.

Along the evaluated ETTh1 Triangular assembly path, lower boundary
discontinuity coincides with lower LSTM TSTR MAE, whereas ARIMA TSTR MAE
increases with $\Delta \mathrm{ACF}_{24h}$. The two panels show that
different forecasters respond to different assembly characteristics,
supporting the joint interpretation of continuity, temporal fidelity, and
downstream forecasting utility.

\FloatBarrier

\subsection{Seed Variability Summary}
\label{app:seed_variability}

To characterize variability across independently trained generator seeds,
we report endpoint summaries for ETTh1 and ETTm1. Each generator-seed-level value is
the mean of 20 sequence replica evaluations. The reported
values are mean $\pm$ sample standard deviation across the three seed-level
averages (seeds 65, 66, and 67), following Section~\ref{sec:metrics}.
Tri-high denotes the highest evaluated triangular-overlap setting: 85.7\%
on ETTh1 and 87.5\% on ETTm1.

\begin{table}[htbp]
\centering
\caption{ETT endpoint variability. Values are mean
$\pm$ sample standard deviation across three independently trained
generator seeds, after averaging 20 sequence replicas within each
seed.}
\label{tab:seed_results_appendix}
\small
\begin{adjustbox}{max width=\linewidth}
\begin{tabular}{llcccc}
\toprule
\multirow{2}{*}{Dataset} &
\multirow{2}{*}{Setting} &
\multicolumn{2}{c}{Diagnostics} &
\multicolumn{2}{c}{TSTR MAE} \\
\cmidrule(lr){3-4}
\cmidrule(lr){5-6}
& &
$J_{\mathrm{jump}}$ &
$\Delta\mathrm{ACF}_{24h}$ &
LSTM & ARIMA \\
\midrule

ETTh1 & Naive &
$4.7745\pm0.2961$ &
$0.0902\pm0.0108$ &
$2.175\pm0.148$ &
$4.578\pm0.189$ \\

ETTh1 & Tri-high &
$0.3964\pm0.0124$ &
$0.1520\pm0.0187$ &
$1.270\pm0.042$ &
$7.386\pm0.369$ \\

ETTm1 & Naive &
$4.1697\pm0.1824$ &
$0.0363\pm0.0062$ &
$3.998\pm0.199$ &
$1.513\pm0.110$ \\

ETTm1 & Tri-high &
$0.1412\pm0.0058$ &
$0.0986\pm0.0180$ &
$1.641\pm0.124$ &
$3.242\pm0.218$ \\

\bottomrule
\end{tabular}
\end{adjustbox}
\end{table}

These seed-level summaries retain the contrast in the corresponding
seed-averaged endpoint results: high-overlap triangular assembly has
lower boundary discontinuity, while LSTM and ARIMA show different downstream responses.

\subsection{Complete Matched TRTR References}
\label{app:complete_trtr}

\tabref{tab:complete_trtr} reports the matched Train-on-Real-Test-on-Real
(TRTR) references for all five downstream forecasters across the four public
evaluation datasets. These references use the corresponding real-data
training protocol and the same target, error scale, and real test segment as
the TSTR evaluations. Lower MAE is better.

\begin{table}[htbp]
\centering
\caption{Matched TRTR MAE for all five forecasters across the four public
evaluation datasets. Values are the reported matched real-training reference
MAE; lower is better.}
\label{tab:complete_trtr}
\small
\begin{adjustbox}{max width=\linewidth}
\begin{tabular}{lccccc}
\toprule
Dataset & LSTM & RNN & LightGBM & ARIMA & Prophet \\
\midrule
ETTh1 & 1.291 & 1.408 & 0.565 & 0.702 & 2.814 \\
ETTm1 & 0.768 & 1.049 & 1.007 & 0.704 & 2.715 \\
Weather & 4.199 & 5.232 & 6.067 & 3.460 & 8.827 \\
Appliances & 27.760 & 29.940 & 30.270 & 23.490 & 32.010 \\
\bottomrule
\end{tabular}
\end{adjustbox}
\end{table}
\FloatBarrier

\section{Complete Sequence Assembly Results for the Main Public Datasets}
\label{app:full_results}
\label{app:benchmark_results}

This section reports the full assembly grid with values averaged first over 20 replicas within each generator seed and then over the three seed-level means. Each setting is evaluated using 20 sequence replicas per checkpoint, with separate downstream training and the same fixed real test set; the reporting hierarchy is described in Section~\ref{sec:metrics}. All tables report the 24-hour diagnostic as $\Delta\mathrm{ACF}_{24h}$. The physical 24-hour lag corresponds to 24 samples for hourly ETTh1, 96 samples for 15-minute ETTm1, and 144 samples for 10-minute Weather and Appliances. For each dataset, we report all evaluated overlap rates, windowing choices, continuity metrics, fidelity metrics, and downstream TSTR MAE values averaged across random seeds. The seed-level summaries are reported in \tabref{tab:seed_results_appendix}. In the tables, \textit{None} denotes no explicit windowing, while \textit{Triangular} and \textit{Hann} denote the corresponding window-weighting choices. The selected configurations in
\figref{fig:assembly_preferences}
are computed from the seed-averaged downstream MAE values reported here. The $0\%$ overlap setting corresponds to naive concatenation without explicit windowing. Consistent with the evaluated grid, the full-grid tables omit $0\%$ Hann and $0\%$ triangular settings.

\newpage
\subsection{ETTh1}

\begin{table}[htbp]
    \centering
    \caption{Complete ETTh1 assembly sweep. Values are means across
    three generator seeds after within-seed replica averaging
    (Section~\ref{sec:metrics}). Lower is better for all reported quantities.}
    \label{tab:etth1_full_sweep}
    \scriptsize
    \setlength{\tabcolsep}{3pt}
    \begin{adjustbox}{max width=\linewidth}
    \begin{tabular}{llcccccccc}
        \toprule
        \multicolumn{1}{c}{\multirow{2}{*}{Window}} &
        \multicolumn{1}{c}{\multirow{2}{*}{Overlap}} &
        \multicolumn{3}{c}{Diagnostics} &
        \multicolumn{5}{c}{TSTR MAE} \\
        \cmidrule(lr){3-5}
        \cmidrule(lr){6-10}
        & & $J_{\mathrm{jump}}$ & Wasserstein &
        $\Delta\mathrm{ACF}_{24h}$ & LSTM & RNN & LightGBM & ARIMA & Prophet \\
        \midrule
        None & 0\% & 4.7745 & 1.7020 & 0.0902 & 2.175 & 2.020 & 1.467 & 4.578 & 5.127 \\
        None & 25\% & 3.0296 & 1.6621 & 0.1076 & 1.588 & 1.864 & 1.475 & 5.369 & 6.185 \\
        Triangular & 25\% & 0.6007 & 1.6847 & 0.1142 & 1.408 & 1.898 & 1.211 & 5.282 & 5.784 \\
        Hann & 25\% & 0.6030 & 1.6945 & 0.0962 & 1.445 & 1.914 & 1.262 & 5.307 & 5.832 \\
        None & 50\% & 2.3990 & 1.6083 & 0.1188 & 1.516 & 1.982 & 1.344 & 6.485 & 6.397 \\
        Triangular & 50\% & 0.5482 & 1.5892 & 0.1277 & 1.348 & 1.804 & 1.312 & 6.004 & 6.065 \\
        Hann & 50\% & 0.5457 & 1.5911 & 0.1085 & 1.435 & 1.482 & 1.561 & 5.852 & 6.015 \\
        None & 75\% & 1.8171 & 1.6642 & 0.1369 & 1.470 & 1.613 & 1.403 & 6.746 & 6.654 \\
        Triangular & 75\% & 0.5289 & 1.5915 & 0.1417 & 1.305 & 1.686 & 1.414 & 6.391 & 6.377 \\
        Hann & 75\% & 0.5302 & 1.5691 & 0.1274 & 1.391 & 1.631 & 1.373 & 6.239 & 6.254 \\
        None & 85.7\% & 1.2056 & 1.7222 & 0.1401 & 1.399 & 1.694 & 1.607 & 7.770 & 6.797 \\
        Triangular & 85.7\% & 0.3964 & 1.6805 & 0.1520 & 1.270 & 1.538 & 1.239 & 7.386 & 6.591 \\
        Hann & 85.7\% & 0.4883 & 1.6464 & 0.1416 & 1.386 & 1.384 & 1.286 & 7.281 & 6.407 \\
        \bottomrule
    \end{tabular}
    \end{adjustbox}
\end{table}

\FloatBarrier

\subsection{ETTm1}

\begin{table}[H]
    \centering
    \caption{Complete ETTm1 assembly sweep. Values are means across
    three generator seeds after within-seed replica averaging
    (Section~\ref{sec:metrics}). Lower is better for all reported quantities.}
    \label{tab:ettm1_full_sweep}
    \scriptsize
    \setlength{\tabcolsep}{3pt}
    \begin{adjustbox}{max width=\linewidth}
    \begin{tabular}{llcccccccc}
        \toprule
        \multicolumn{1}{c}{\multirow{2}{*}{Window}} &
        \multicolumn{1}{c}{\multirow{2}{*}{Overlap}} &
        \multicolumn{3}{c}{Diagnostics} &
        \multicolumn{5}{c}{TSTR MAE} \\
        \cmidrule(lr){3-5}
        \cmidrule(lr){6-10}
        & & $J_{\mathrm{jump}}$ & Wasserstein &
        $\Delta\mathrm{ACF}_{24h}$ & LSTM & RNN & LightGBM & ARIMA & Prophet \\
        \midrule
        None & 0\% & 4.1697 & 1.6681 & 0.0363 & 3.998 & 4.926 & 2.602 & 1.513 & 1.270 \\
        None & 25\% & 2.4517 & 1.5843 & 0.0550 & 3.855 & 6.549 & 1.224 & 1.740 & 1.340 \\
        Triangular & 25\% & 0.2766 & 1.6681 & 0.0793 & 3.209 & 7.521 & 0.889 & 1.801 & 1.369 \\
        Hann & 25\% & 0.2699 & 1.7062 & 0.0702 & 2.636 & 5.768 & 0.958 & 1.755 & 1.366 \\
        None & 50\% & 2.1813 & 1.4130 & 0.0503 & 3.351 & 6.344 & 1.566 & 1.884 & 1.763 \\
        Triangular & 50\% & 0.2262 & 1.5305 & 0.0791 & 3.925 & 5.876 & 1.001 & 1.776 & 1.412 \\
        Hann & 50\% & 0.2272 & 1.5907 & 0.0747 & 2.565 & 5.319 & 1.030 & 1.690 & 1.397 \\
        None & 75\% & 1.2815 & 1.0852 & 0.0785 & 3.158 & 3.695 & 0.964 & 2.878 & 2.255 \\
        Triangular & 75\% & 0.2274 & 1.1726 & 0.0927 & 2.986 & 4.785 & 0.918 & 2.704 & 2.256 \\
        Hann & 75\% & 0.2259 & 1.2113 & 0.0900 & 3.028 & 3.227 & 0.952 & 2.591 & 2.217 \\
        None & 87.5\% & 0.6912 & 1.0054 & 0.0915 & 2.409 & 4.814 & 1.011 & 3.372 & 2.448 \\
        Triangular & 87.5\% & 0.1412 & 1.0624 & 0.0986 & 1.641 & 3.945 & 0.911 & 3.242 & 2.472 \\
        Hann & 87.5\% & 0.1412 & 1.0838 & 0.0967 & 2.291 & 3.590 & 0.922 & 3.145 & 2.437 \\
        \bottomrule
    \end{tabular}
    \end{adjustbox}
\end{table}

\FloatBarrier

\subsection{Weather}
\label{app:weather_results}

\tabref{tab:weather_full_sweep} reports the mean results for all 13
Weather assembly configurations: one zero-overlap None baseline
and 12 positive-overlap settings. This sweep places the
endpoint comparison in the main text within the full assembly grid
and
windowing choices. Results follow the aggregation protocol in
Section~\ref{sec:metrics}, with a fixed diffusion checkpoint within each
generator seed.

\begin{table}[htbp]
    \centering
    \caption{Complete Weather assembly sweep. Values are means across
    three generator seeds after averaging 20 sequence replicas within
    each seed. Lower is better for all metrics.}
    \label{tab:weather_full_sweep}
    \scriptsize
    \setlength{\tabcolsep}{3pt}
    \begin{adjustbox}{max width=\linewidth}
    \begin{tabular}{llcccccccc}
        \toprule
        \multicolumn{1}{c}{\multirow{2}{*}{Window}} &
        \multicolumn{1}{c}{\multirow{2}{*}{Overlap}} &
        \multicolumn{3}{c}{Diagnostics} &
        \multicolumn{5}{c}{TSTR MAE} \\
        \cmidrule(lr){3-5}
        \cmidrule(lr){6-10}
        & & $J_{\mathrm{jump}}$ & Wasserstein &
        $\Delta\mathrm{ACF}_{24h}$ & LSTM & RNN & LightGBM & ARIMA & Prophet \\
        \midrule
        None & 0\% & 47.711 & 56.830 & 0.2312 & 11.518 & 20.095 & 9.397 & 15.767 & 17.462 \\
        None & 25\% & 37.722 & 56.193 & 0.2733 & 10.124 & 18.811 & 11.728 & 17.644 & 17.500 \\
        Triangular & 25\% & 39.253 & 56.197 & 0.2761 & 10.463 & 18.768 & 10.092 & 17.486 & 17.210 \\
        Hann & 25\% & 38.940 & 56.187 & 0.2662 & 10.225 & 18.336 & 10.822 & 17.126 & 15.972 \\
        None & 50\% & 34.463 & 58.700 & 0.2347 & 13.085 & 15.724 & 10.786 & 20.517 & 17.607 \\
        Triangular & 50\% & 34.862 & 57.685 & 0.2482 & 9.471 & 17.824 & 11.014 & 19.164 & 17.365 \\
        Hann & 50\% & 33.469 & 57.538 & 0.2443 & 8.530 & 16.887 & 10.898 & 18.457 & 16.237 \\
        None & 75\% & 32.988 & 57.887 & 0.2734 & 9.672 & 16.517 & 9.868 & 25.531 & 19.658 \\
        Triangular & 75\% & 31.523 & 57.601 & 0.2795 & 9.603 & 19.030 & 8.905 & 23.752 & 17.462 \\
        Hann & 75\% & 31.087 & 57.402 & 0.2690 & 9.411 & 15.242 & 9.797 & 22.527 & 16.537 \\
        None & 85.7\% & 31.541 & 55.514 & 0.3559 & 9.148 & 15.155 & 7.907 & 27.972 & 19.834 \\
        Triangular & 85.7\% & 29.853 & 56.803 & 0.3378 & 8.752 & 14.613 & 9.657 & 27.282 & 17.541 \\
        Hann & 85.7\% & 30.667 & 56.999 & 0.3144 & 8.244 & 14.540 & 10.575 & 26.228 & 16.610 \\
        \bottomrule
    \end{tabular}
    \end{adjustbox}
\end{table}

The full sweep provides three main results. First, all valid positive-overlap
settings have lower reported
$J_{\mathrm{jump}}$ than the zero-overlap baseline, and within each
evaluated weighting family, $J_{\mathrm{jump}}$ decreases as overlap increases.
Second, LSTM MAE is non-monotonic and reaches its lowest reported mean at 85.7\%
Hann overlap (8.244), whereas the lowest reported
$J_{\mathrm{jump}}$ occurs at 85.7\% triangular overlap (29.853). Third, every
positive-overlap setting has higher reported
$\Delta\mathrm{ACF}_{24h}$ and higher ARIMA MAE than the zero-overlap
baseline. The full Weather grid therefore extends the endpoint comparison across the
evaluated overlap range.

\FloatBarrier

\subsection{Appliances}
\label{app:appliances_results}

\tabref{tab:appliances_full_sweep} reports the complete Appliances
assembly sweep. Values use the same seed and replica aggregation as
the Weather results and retain the dataset-specific 85.7\% maximum overlap.

\begin{table}[htbp]
    \centering
    \caption{Complete Appliances assembly sweep. Values are means
    across three generator seeds after averaging 20 sequence replicas
    within each seed.}
    \label{tab:appliances_full_sweep}
    \scriptsize
    \setlength{\tabcolsep}{3pt}
    \begin{adjustbox}{max width=\linewidth}
    \begin{tabular}{llcccccccc}
        \toprule
        \multicolumn{1}{c}{\multirow{2}{*}{Window}} &
        \multicolumn{1}{c}{\multirow{2}{*}{Overlap}} &
        \multicolumn{3}{c}{Diagnostics} &
        \multicolumn{5}{c}{TSTR MAE} \\
        \cmidrule(lr){3-5}
        \cmidrule(lr){6-10}
        & & $J_{\mathrm{jump}}$ & Wasserstein &
        $\Delta\mathrm{ACF}_{24h}$ & LSTM & RNN & LightGBM & ARIMA & Prophet \\
        \midrule
        None & 0\% & 216.001 & 29.729 & 0.0755 & 48.600 & 53.050 & 52.060 & 50.090 & 45.060 \\
        None & 25\% & 160.422 & 29.796 & 0.0692 & 43.600 & 50.940 & 49.900 & 56.910 & 45.810 \\
        Triangular & 25\% & 109.579 & 29.427 & 0.0681 & 45.860 & 55.130 & 52.270 & 56.840 & 44.530 \\
        Hann & 25\% & 106.337 & 29.277 & 0.0679 & 47.780 & 52.740 & 52.640 & 56.880 & 44.190 \\
        None & 50\% & 142.913 & 32.194 & 0.0721 & 42.280 & 50.870 & 51.820 & 56.880 & 58.410 \\
        Triangular & 50\% & 97.655 & 31.079 & 0.0705 & 42.970 & 55.650 & 51.350 & 53.090 & 59.310 \\
        Hann & 50\% & 97.119 & 30.858 & 0.0700 & 42.940 & 54.700 & 52.030 & 52.130 & 58.930 \\
        None & 75\% & 129.892 & 33.438 & 0.0651 & 40.580 & 50.500 & 49.580 & 71.100 & 75.640 \\
        Triangular & 75\% & 95.634 & 32.052 & 0.0672 & 39.960 & 51.820 & 47.860 & 66.820 & 74.410 \\
        Hann & 75\% & 95.572 & 31.776 & 0.0675 & 40.920 & 51.870 & 49.120 & 65.270 & 73.490 \\
        None & 85.7\% & 120.096 & 34.819 & 0.0685 & 39.820 & 49.460 & 47.540 & 67.080 & 69.850 \\
        Triangular & 85.7\% & 94.367 & 33.598 & 0.0692 & 38.630 & 50.790 & 48.500 & 60.970 & 69.010 \\
        Hann & 85.7\% & 94.502 & 33.334 & 0.0698 & 39.770 & 50.180 & 51.050 & 60.430 & 67.970 \\
        \bottomrule
    \end{tabular}
    \end{adjustbox}
\end{table}

For Appliances, the lowest measured junction discontinuity and the lowest
LSTM MAE both occur at the 85.7\% triangular setting, whereas other
forecasters favor different configurations, further illustrating
forecaster-dependent assembly preferences.

\section{Operational Electricity-Load Application}
\label{app:public_operational_results}

\subsection{Operational Protocol}
The operational-load series provides the domain-specific application and validation setting for the forecaster-dependent assembly regularities identified on the four public datasets. It uses hourly
sampling and an aggregate-load target. The chronological training source
interval is 2023-01-01 01:00--2023-03-12 00:00 and the real test interval is
2023-03-12 01:00--2023-03-13 00:00, giving a 24-hour real evaluation window.
The documented operational DP-SGD noise multiplier is $\sigma=2.041$ and
the calibration release uses raw clipping bound $B=6\times10^{7}$ with the
calibration parameters in \hyperref[app:dp_details]{Appendix~B}
(including $\epsilon_{\mathrm{cal}}=1$,
$\delta_{\mathrm{cal}}=10^{-6}$, $\alpha=0.7$, seed 0, $\kappa=5$, and
smoothing window 3). The seed and replica protocol is the same documented
three generator seeds with 20 sequence replicas per seed.

\subsection{Complete Sequence Assembly Results}
\label{app:operational_grid}

The table below reports all 13 valid assembly configurations for
the operational electricity-load case. Values are means
of three generator-seed-level averages, with 20 sequence replicas
averaged within each seed. Bold values identify the minimum
value within the evaluated grid for the corresponding diagnostic or
forecaster.

\begin{table}[htbp]
    \centering
    \caption{Complete operational electricity-load assembly grid.
    Values are means across three generator-seed-level averages,
    with 20 sequence replicas averaged within each seed. Lower is better
    for all reported quantities.}
    \label{tab:operational_results}
    \scriptsize
    \setlength{\tabcolsep}{3pt}
    \begin{adjustbox}{max width=\linewidth}
    \begin{tabular}{llcccccccc}
        \toprule
        \multicolumn{1}{c}{\multirow{2}{*}{Window}} &
        \multicolumn{1}{c}{\multirow{2}{*}{Overlap}} &
        \multicolumn{3}{c}{Diagnostics} &
        \multicolumn{5}{c}{TSTR MAE} \\
        \cmidrule(lr){3-5}
        \cmidrule(lr){6-10}
        & & $J_{\mathrm{jump}}$ & Wasserstein &
        $\Delta\mathrm{ACF}_{24h}$ & LSTM & RNN & LightGBM & ARIMA & Prophet \\
        \midrule
        None & 0\% & 3.892 & 1.110 & 0.0439 & 1.685 & 1.961 & \textbf{0.900} & 2.667 & \textbf{2.220} \\
        None & 25\% & 1.986 & 1.094 & 0.0658 & 1.573 & 1.812 & 0.959 & 2.611 & 2.353 \\
        Triangular & 25\% & 1.207 & 1.157 & 0.0741 & 1.724 & 1.941 & 0.946 & 2.696 & 2.246 \\
        Hann & 25\% & 1.072 & 1.193 & 0.0411 & 1.355 & 2.645 & 0.930 & \textbf{2.599} & 2.301 \\
        None & 50\% & 1.528 & 0.656 & 0.0463 & 1.321 & 2.374 & 1.111 & 2.763 & 2.670 \\
        Triangular & 50\% & 1.121 & 0.838 & 0.0670 & 1.551 & 2.152 & 1.036 & 2.763 & 2.355 \\
        Hann & 50\% & 1.045 & 0.949 & 0.0408 & 1.314 & 2.303 & 1.204 & 2.921 & 2.319 \\
        None & 75\% & 1.286 & 0.384 & 0.0676 & \textbf{1.147} & 2.499 & 1.435 & 2.900 & 2.765 \\
        Triangular & 75\% & 0.862 & 0.488 & 0.0976 & 1.409 & 2.454 & 1.150 & 3.082 & 2.609 \\
        Hann & 75\% & 0.816 & 0.561 & 0.0378 & 1.287 & 2.001 & 1.223 & 3.092 & 2.509 \\
        None & 85.7\% & 0.996 & \textbf{0.372} & 0.0867 & 1.308 & 1.996 & 1.016 & 2.870 & 3.148 \\
        Triangular & 85.7\% & 0.815 & 0.405 & 0.1167 & 1.365 & \textbf{1.640} & 0.963 & 2.835 & 2.811 \\
        Hann & 85.7\% & \textbf{0.584} & 0.429 & \textbf{0.0346} & 1.366 & 1.960 & 0.979 & 2.858 & 2.799 \\
        \bottomrule
    \end{tabular}
    \end{adjustbox}
\end{table}

\section{Cross-Generator Check with DP-ConvVAE}
\label{app:vae_robustness}
\label{app:generator_comparison}

\subsection{Protocol and Sequence Assembly Settings}

The DP-ConvVAE experiment retains the ETT target, preprocessing,
privacy-budget scope, calibration, and downstream evaluation protocols.
The target is OT. ETTh is sampled hourly and uses
$(L,T_{\mathrm{final}})=(168,1680)$; ETTm is sampled at
15-minute intervals and uses $(672,6720)$. Both families therefore
use seven-day generated windows and 70-day assembled training
sequences. The real evaluation window spans 24 hours, corresponding to 24 samples for
ETTh and 96 samples for ETTm.
Three generators are trained independently
with different seeds for each dataset. Each generator supplies 20
sequence replicas per setting, constructed from
independently sampled generated windows. LSTM and ARIMA are fitted
separately to each replica and evaluated on the same fixed real test
set. Within each dataset and generator seed, $C0$, $C1$, and $C2$
draw from a shared generated-window pool.

The training and calibration parameters are
$\epsilon_{\mathrm{train}}=9$, $\delta_{\mathrm{train}}=10^{-5}$
and $\epsilon_{\mathrm{cal}}=1$, $\delta_{\mathrm{cal}}=10^{-6}$,
respectively, with the separate privacy units described in
Section~\ref{sec:privacy_scope}.

\begin{table}[htbp]
\centering
\caption{DP-ConvVAE assembly settings with seven-day windows
and 70-day training sequences. $C1$ and $C2$ share the same stride
and overlap within each dataset family.}
\label{tab:vae_protocol}
\small
\begin{adjustbox}{max width=\linewidth}
\begin{tabular}{llccccl}
\toprule
Dataset family & Setting & $L$ & $T_{\mathrm{final}}$ & Stride $k$ & Overlap & Weighting \\
\midrule
ETTh & $C0$ & 168 & 1680 & 168 & 0\% & None (concatenation) \\
ETTh & $C1$ & 168 & 1680 & 24 & 85.7\% & Equal weights \\
ETTh & $C2$ & 168 & 1680 & 24 & 85.7\% & Triangular \\
\midrule
ETTm & $C0$ & 672 & 6720 & 672 & 0\% & None (concatenation) \\
ETTm & $C1$ & 672 & 6720 & 84 & 87.5\% & Equal weights \\
ETTm & $C2$ & 672 & 6720 & 84 & 87.5\% & Triangular \\
\bottomrule
\end{tabular}
\end{adjustbox}
\end{table}

$C0$ directly concatenates generated windows. $C1$ averages their
contributions in overlapping regions using normalized equal weights.
$C2$ applies normalized triangular weights, reducing the contribution
of window edges. The $C1$--$C2$ comparison evaluates weighting choices
at a fixed overlap, while $C0$ supplies the zero-overlap reference.

\subsection{Repetitions and Metric Aggregation}

The five reported metrics are $J_{\mathrm{jump}}$, Wasserstein distance,
24-hour autocorrelation deviation, LSTM MAE, and ARIMA MAE, following the
definitions in Sections~\ref{sec:principle_boundary}
and~\ref{sec:joint_evaluation}.
The 24-hour lag corresponds to 24 samples for ETTh and 96 for ETTm.

The aggregation hierarchy is identical to that of the primary
diffusion-generator experiments.
For each metric $q$ and assembly setting, let $q_{g,r}$ denote
the result for generator seed $g$ and replica $r$. The seed-level
mean and final reported mean are
\[
    \bar{q}_g=\frac{1}{20}\sum_{r=1}^{20}q_{g,r},
    \qquad
    \bar{q}=\frac{1}{3}\sum_{g=1}^{3}\bar{q}_g.
\]
The reported sample standard deviation is computed across the three
seed-level means as
\[
    s_q=\sqrt{\frac{1}{3-1}\sum_{g=1}^{3}(\bar{q}_g-\bar{q})^2}.
\]
Thus, each dataset, setting, and forecaster contributes
60 replica-level evaluations nested within three independently trained
generators. The uncertainty summaries describe variation across the
generator-seed-level averages.

\subsection{Five-Metric Results and Interpretation}

\tabref{tab:vae_full_results} organizes the two datasets and three
settings in mean $\pm$ standard deviation form. Each cell summarizes
three generator-seed-level averages, each computed from 20
sequence replicas.

\begin{table}[htbp]
\centering
\caption{DP-ConvVAE results across ETTh1 and ETTm1 and three
assembly settings. For each generator seed, results are first
averaged over 20 sequence replicas. Cells report mean $\pm$ sample
standard deviation across the three seed-level means from independently
trained generators. Lower is better for all five metrics.}
\label{tab:vae_full_results}
\scriptsize
\setlength{\tabcolsep}{3pt}
\begin{adjustbox}{max width=\linewidth}
\begin{tabular}{llccccc}
\toprule
Dataset & Setting & $J_{\mathrm{jump}}$ & Wasserstein &
$\Delta\mathrm{ACF}_{24h}$ & LSTM MAE & ARIMA MAE \\
\midrule
ETTh1 & C0 & $0.6892\pm0.0446$ & $2.7101\pm0.1433$ & $0.3405\pm0.0434$ & $3.080\pm0.242$ & $2.952\pm0.193$ \\
ETTh1 & C1 & $0.1233\pm0.0040$ & $2.8880\pm0.1804$ & $0.0410\pm0.0069$ & $2.947\pm0.096$ & $3.215\pm0.140$ \\
ETTh1 & C2 & $0.0549\pm0.0020$ & $2.8716\pm0.1208$ & $0.0183\pm0.0025$ & $2.814\pm0.201$ & $4.154\pm0.205$ \\
\midrule
ETTm1 & C0 & $0.4812\pm0.0252$ & $1.6217\pm0.1057$ & $0.8623\pm0.1542$ & $2.848\pm0.227$ & $4.020\pm0.269$ \\
ETTm1 & C1 & $0.2125\pm0.0125$ & $1.7561\pm0.0864$ & $0.9312\pm0.1357$ & $2.822\pm0.140$ & $4.225\pm0.280$ \\
ETTm1 & C2 & $0.1770\pm0.0070$ & $1.7475\pm0.1141$ & $0.9271\pm0.1599$ & $2.776\pm0.196$ & $4.319\pm0.221$ \\
\bottomrule
\end{tabular}
\end{adjustbox}
\end{table}

Across the two ETT datasets, the reported mean $J_{\mathrm{jump}}$
decreases from $C0$ through $C1$ to $C2$, accompanied by decreasing
LSTM MAE. ARIMA MAE increases across the three settings on ETTh1 and
ETTm1. The second generator exhibits the same forecaster-dependent contrast
between boundary continuity and downstream forecasting response.

%% HARV author--year bibliography, matching the Overleaf template.
\section*{Declaration of generative AI and AI-assisted technologies in the manuscript preparation process}

During the preparation of this work, the authors used ChatGPT (OpenAI) for language editing and improving clarity. After using this tool, the authors reviewed and edited the content as needed and take full responsibility for the content of the publication.

%% References are embedded directly in this .tex file for submission portability.

\end{document}